\documentclass{article}

\PassOptionsToPackage{numbers, compress}{natbib}
 \usepackage[preprint]{neurips_2024}

\usepackage{enumerate,letltxmacro}
\LetLtxMacro\itemold\item
\usepackage{enumitem}

\usepackage[utf8]{inputenc} % allow utf-8 input
\usepackage[T1]{fontenc}    % use 8-bit T1 fonts
\usepackage{hyperref}       % hyperlinks
\usepackage{url}            % simple URL typesetting
\usepackage{booktabs}       % professional-quality tables
\usepackage{amsfonts}       % blackboard math symbols
\usepackage{nicefrac}       % compact symbols for 1/2, etc.
\usepackage{microtype}      % microtypography
\usepackage{xcolor}         % colors

\usepackage{graphicx}
\usepackage{subfig}
\usepackage{amsmath}   
\usepackage[capitalize]{cleveref}
\usepackage{colortbl}
\usepackage{multirow}
\usepackage{listings}
\usepackage{xcolor}
\usepackage{tcolorbox}
\usepackage{tabularx}
\usepackage{caption}
\usepackage{fontawesome}
\tcbuselibrary{most}
\tcbuselibrary{skins}
\usepackage{makecell}
\usepackage{wrapfig,lipsum,booktabs}

\crefname{section}{Sec.}{Secs.}
\Crefname{section}{Section}{Sections}
\Crefname{table}{Table}{Tables}
\crefname{table}{Tab.}{Tabs.}
\crefname{figure}{Fig.}{Figs.}
\Crefname{figure}{Figure}{Figures}
\crefname{appendix}{Appx.}{Appxs.}
\Crefname{appendix}{Appendix}{Appendixs}
\crefname{subsection}{Sec.}{Secs.}

\newcommand\nnfootnote[1]{%
  \begin{NoHyper}
  \renewcommand\thefootnote{}\footnotetext{#1}%
  \addtocounter{footnote}{-1}%
  \end{NoHyper}
}
\newcommand\ntfootnote[1]{%
  \begin{NoHyper}
  \renewcommand\thefootnote{}\footnotetext{#1}%
  \addtocounter{footnote}{0}%
  \end{NoHyper}
}

\newcommand{\tablestyle}[2]{\setlength{\tabcolsep}{#1}\renewcommand{\arraystretch}{#2}\centering\footnotesize}
\newlength\savewidth

\definecolor{maroon}{cmyk}{0,0.1,0.01,0.01}
\definecolor{blue}{cmyk}{0.95,0.0,0.2,0.2}
\definecolor{yellow}{cmyk}{0.01,0.0,0.2,0.01}
\definecolor{lightblue}{cmyk}{0.1,0.0,0.02,0.02}
\definecolor{case_verb}{HTML}{fbde84}
\definecolor{case_adj}{HTML}{cccdff}
\definecolor{case_noun}{HTML}{bfeaf1}
\definecolor{case_ff}{HTML}{e65352}
\definecolor{case_error}{HTML}{ffff00}
\definecolor{darkgreen}{RGB}{51,181,41}
\definecolor{darkorange}{RGB}{252,135,62}
\definecolor{t_green}{HTML}{f1f2e4}

\definecolor{lightgray}{gray}{0.95}
\lstdefinestyle{prompt}{
    basicstyle=\ttfamily\fontsize{7pt}{8pt}\selectfont,
    frame=none,
    breaklines=true,
    backgroundcolor=\color{lightgray},
    breakatwhitespace=true,
    breakindent=0pt,
    escapeinside={(*@}{@*)},
    numbers=none,
    numbersep=5pt,
    xleftmargin=5pt,
}
\tcbset{
  aibox/.style={
    top=10pt,
    colback=white,
    colframe=black,
    colbacktitle=black,
    enhanced,
    center,
    attach boxed title to top left={yshift=-0.1in,xshift=0.15in},
    boxed title style={boxrule=0pt,colframe=white,},
  }
}
\newtcolorbox{AIbox}[2][]{aibox, title=#2,#1}

\title{Prism: A Framework for Decoupling and Assessing the Capabilities of VLMs}

\author{%
  Yuxuan Qiao$^{2\ast}$, Haodong Duan$^{1\dag}$, Xinyu Fang$^{4}$, Junming Yang$^{5}$, Lin Chen$^{6}$,  \\
  \textbf{Songyang Zhang$^{1}$, Jiaqi Wang$^{1}$, Dahua Lin$^{1,3}$, Kai Chen$^{1\dag}$} \\
  $^{1}$Shanghai AI Laboratory \quad $^{2}$Nanjing University \quad $^{3}$The Chinese University of Hong Kong\\
  $^{4}$Tongji University \quad $^{5}$Nanjing University of Posts and Telecommunications\\
  $^{6}$University of Science and Technology of China \\
  \texttt{yuxuanqiao@smail.nju.edu.cn} \\
  \texttt{duanhaodong@pjlab.org.cn}
}

\begin{document}
\maketitle
\nnfootnote{\noindent $^{\ast}$ The work was done during an internship at Shanghai AI Laboratory.}
\ntfootnote{$^{\dag}$ Corresponding Author.}

\begin{abstract}

Vision Language Models (VLMs) demonstrate remarkable proficiency in addressing a wide array of visual questions, 
which requires strong perception and reasoning faculties. 
Assessing these two competencies independently is crucial for model refinement, 
despite the inherent difficulty due to the intertwined nature of seeing and reasoning in existing VLMs. 
To tackle this issue, we present \textbf{Prism}, 
an innovative framework designed to disentangle the perception and reasoning processes involved in visual question solving.
Prism comprises two distinct stages:
a \textbf{perception stage} that utilizes a VLM to extract and articulate visual information in textual form, 
and a \textbf{reasoning stage} that formulates responses based on the extracted visual information using a Large Language Model (LLM).
This modular design enables the systematic comparison and assessment of both proprietary and open-source VLM for their perception and reasoning strengths.
Our analytical framework provides several valuable insights, 
underscoring Prism's potential as a cost-effective solution for vision-language tasks.
By combining a streamlined VLM focused on perception with a powerful LLM tailored for reasoning,
Prism achieves superior results in general vision-language tasks 
while substantially cutting down on training and operational expenses.
Quantitative evaluations show that Prism, 
when configured with a vanilla 2B LLaVA and freely accessible GPT-3.5, 
delivers performance on par with VLMs $10 \times$ larger on the rigorous multimodal benchmark MMStar.
The project is released at: \url{https://github.com/SparksJoe/Prism}.

\end{abstract}

\section{Introduction}
% 近期，得益于LLM的长足进步，LVLM的发展得到了推动，缩减着我们与通用人工智能的距离。诸如GPT4V, GeminiProVision, QwenVLMax等专用商业模型正在展现着他们出色的通用能力，解决着复杂的多模态任务。同时，开源领域涌现出的模型也在某些任务上与上述闭源模型取得了相当的成绩，如LLaVA_Next。他们主要采取将视觉输入作为token处理的策略，通过image-text对比学习的vision encoder (e.g CLIP) 编码为视觉embedding, 并经过projector对齐到文本空间，借助如llama, vicuna, mistral， internlm等LLM的泛化能力进行复杂的视觉语言对话。

With the rapid development of Large Language Models (LLMs) \cite{2022chatgpt,touvron2023llama,touvron2023llama2,2023internlm,qwen,du2022glm,Claude3}, 
Vision Language Models (VLMs) \cite{OpenAI2023GPT4TR,team2023gemini,bai2023qwen,liu2023visual,dai2023instructblip,internlmxcomposer2} have also experienced significant advancements. 
As an end-to-end approach, VLMs trained on large-scale multimodal data
\cite{liu2023visual,chen2023sharegpt4v,schuhmann2021laion,changpinyo2021conceptual} exhibit superior performance on a variety of tasks. 
These tasks range from basic ones such as object localization~\cite{lin2014microsoft} and optical character recognition~\cite{mishra2019ocr,liu2023hidden}
to more complex challenges like document or diagram comprehension~\cite{kembhavi2016diagram,mathew2021docvqa,masry2022chartqa} 
and solving geometric problems~\cite{lu2023mathvista}. 
For VLMs to solve a general visual question, 
two essential capabilities are required:
1) \textbf{Perception}: extracting necessary information from the image;
2) \textbf{Reasoning}: generating answers based on the extracted information and contextual understanding. 
Limitations in either capability can impede the overall performance of a VLM.

% Recently, significant advancements in large language models (LLMs) have propelled the development of large vision-language Models (LVLMs), thereby narrowing the gap towards artificial general intelligence (AGI). Specialized commercial models such as GPT-4V, GeminiProVision and QwenVLMax are demonstrating their exceptional general capabilities by tackling complex multimodal tasks. Concurrently, open-source models such as LLaVA-NeXT \cite{liu2024llavanext} are achieving comparable results in certain tasks against these proprietary models. These systems primarily adopt a strategy of processing visual inputs as tokens. They utilize vision encoders (e.g., CLIP) for image-text contrastive learning to generate visual embeddings. These embeddings are then aligned to the textual domain through a projector, leveraging the generalization abilities of LLMs such as Llama, Vicuna, Mistral, and InternLM to facilitate complex vision-language interactions.

% 在VLM中，感知系统负责理解和解析视觉输入，而推理系统则处理逻辑和语言层面的任务。一些近期的工作认为LVLM解决多模态问题主要依靠模型的感知与推理，并且一些工作发现感知和推理都限制着模型的表现。以SOTA模型GPT4V为例，在感知方面，它仍然不能完全处理好对人类来说较为简单的视觉感知，在推理方面，受制于能力的局限，在推理要求较高的专业问题上，它也往往会出错。(our cases)。研究LVLM的感知和推理的重要性在于，确认限制模型的能力维度能够更有效的指引模型迭代方向。

A systematic evaluation of the perception and reasoning capabilities is crucial to provide valuable insights for future model optimization.
However, seeing and reasoning are mostly \textbf{entangled} in existing VLMs.  
Proprietary VLMs, such as GPT-4v are opaque systems that only provide the final answer to visual questions.
Meanwhile, open-source VLMs~\cite{liu2023visual} commonly utilize vision encoders to extract visual embeddings, 
which are often difficult to interpret, 
from the image and employ adapted LLMs to generate answers based on these visual and linguistic embeddings.
In light of this challenge, 
we introduce \textbf{Prism}, a framework designed to disentangle the perception and reasoning processes of any given VLM.
It can serve as a proxy for assessing the real perception capabilities of VLMs. 

% In pursuit of genuine artificial general intelligence, VLMs are expected to exhibit strong perception and reasoning abilities. Perception is tasked with the understanding and parsing of visual inputs, while reasoning encompasses tasks that require logic and linguistic understanding. Additionally, research has revealed that limitations in both perception and reasoning are widely present in the model's performance. The state-of-the-art model GPT-4V, for instance, encounters difficulties with visual perception tasks that humans find relatively straightforward. When it comes to reasoning, the model is particularly susceptible to mistakes in scenarios that demand complex reasoning, especially in specialized fields.

% 不幸的是，系统的分析LVLM的感知能力和推理能力仍然是一个空缺，无法确认限制模型的能力维度以完成更有效的模型迭代。并且在现有的benchmark上，LVLM的能力往往是通过做选择题来展现。这样的主流方式可能会使模型被overfit到选择题上，并且这样的主流方式也无法很好的确认模型的内在能力。为了填补这一部分空缺，我们希望以全新的，系统的方式研究现有LVLM的感知能力与推理能力。

Prism decomposes the process of solving general visual questions into two distinct stages:
1) a \textbf{perception stage} that concentrates on extracting visual information from the image using VLMs and articulating this information in textual form;
and 2) a \textbf{reasoning stage} that utilizes LLMs to answer the question based on the extracted visual information. 
Prism facilitates the analysis of VLMs' capabilities through two approaches. 
To assess the true perception capabilities, 
one can use a constant LLM in the reasoning stage while testing various VLMs in the perception stage.
Conversely, by keeping the VLM fixed and varying the LLM in the reasoning stage, 
one can determine whether a VLM's performance is limited by its reasoning capabilities. 
Through this analysis, we have uncovered several key insights: 
1) Proprietary VLMs, such as GPT-4o and GPT-4v, 
take the lead in the perception capabilities competition; 
2) For open-source VLMs, perception capabilities remain relatively consistent regardless of the language model's size; 
and 3) The overall performance of open-source VLMs, particularly those with smaller-scale language models like 7B variants, is often constrained by the limited reasoning capabilities. 

\begin{figure}[t]
  \centering
  \includegraphics[width=.85\textwidth]{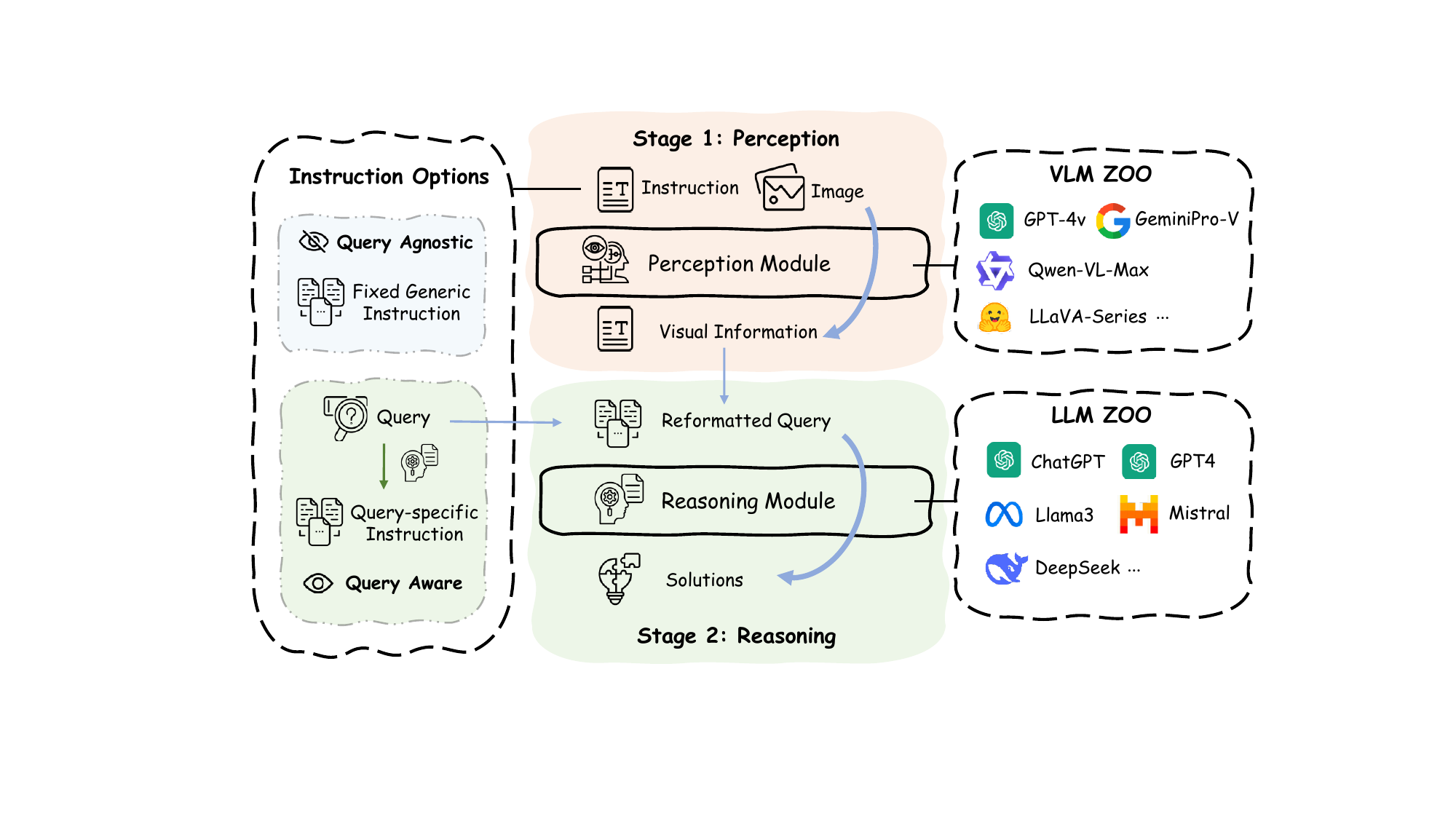}
  \vspace{-1.5mm}
  \caption{\textbf{Prism Framework Architecture.} Prism framework takes image-query pairs as input.
  An instruction (can be query-agnostic or query-aware) and the image are first fed into the VLM to extract visual information.
  Then, an LLM is used to generate the answer based on the reformatted query which combines the original question and visual information in textual form. }
  \label{fig:Prism}
  \vspace{-7mm}
\end{figure}

% \begin{figure*}[ht]
%   \subfloat[End-to-End VLMs]{\label{fig:e2e}\includegraphics[width=0.38\textwidth]{figs/E2E.png}}
%   \hspace{0.01\textwidth}
%   \subfloat[The Prism Framework. For image-query input, the chosen instruction and image are used for VLM to extract visual information and LLM generates answer based on the textual information. The instruction can derive from a query-specific process or be purely query-agnostic.]{\label{fig:prism}\includegraphics[width=0.59\textwidth]{figs/Prism.png}}
%   \hspace{0.01\textwidth}
%   \caption{Comparing the Architecture of End-to-End VLMs and the Prism Framework. }
%   \label{fig:Comparison}
% \end{figure*}

% Unfortunately, there is currently a gap in systematically analyzing the perception and reasoning capabilities of VLMs, which hinders the identification of the dimensions that limit a model's abilities and hence the effective iteration of models. Moreover, the proficiency of VLMs is often assessed through multiple-choice questions in existing benchmarks. Such a prevalent method may lead to models being overfitted to multiple-choice formats and does not adequately capture their intrinsic abilities. 

% 受wittgenstein “The limits of my language means the limits of my world.”的启发，我们提出一套显示的框架，将LVLM的感知和推理解耦，模拟模型解决问题的过程，而LVLM在框架中只负责感知，并使用LLM作为推理工具，整合模型的文本描述解决图片相关的QA。在此基础设定之下，VLM在Prism中只发挥它的感知能力，这允许我们设置相应的推理模型以系统展现模型的感知性能。

Beyond its role as an evaluation framework, 
Prism also excels as an efficient general Vision-Language Model (VLM).
Building upon findings 2 and 3, 
we posit that integrating a small-scale VLM as a visual captioner 
with a powerful LLM as a reasoning engine 
offers a promising and efficient strategy for general vision-language tasks.
By concentrating on visual information extraction, 
a lightweight VLM can achieve decent performance on par with much larger VLMs.
When paired with a powerful yet economical LLM, 
thanks to advancements in deployment techniques,
one can achieve a robust solution for visual-language information processing, 
requiring significantly fewer hardware resources for training and deployment.
In our experiments, we trained an approximately 2B-parameter vanilla LLaVA to extract visual information 
and observed that it exhibits perception performance comparable to LLaVA-NeXT~\cite{liu2024llavanext} that is equipped with a 34B powerful language model. 
Quantitative evaluations indicate that Prism,
when instantiated with a streamlined visual captioner and the freely available ChatGPT-3.5, outperforms many open-source VLMs on multiple multimodal benchmarks including the stringent multimodal understanding benchmark MMStar~\cite{chen2024we}.
Notably, the advantage is particularly pronounced on reasoning-related visual questions.

% Building VLMs from scratch requires huge amounts of visual instruction tuning data and

% To bridge this gap, we propose a novel and systematic framework called Prism, which refracts out the intrinsic abilities within VLMs through decoupling perception and reasoning. Prism simulates an ideal problem-solving process where the VLM is responsible for describing the contents in images, while the reasoning about answering questions based on descriptions is handled as a text-only task assigned to LLM. Under this foundational setting, the VLM operates solely on its perceptual abilities within the Prism framework, allowing for systematically demonstrating the model's perception performance by selectively pairing it with corresponding reasoning models.

% Leveraging Prism, we evaluated extensive open-source models and commercial API models on vision-dependent benchmarks to assess their intrinsic perception capabilities. We found that perception abilities indeed constrain the models' performance, exhibiting a bottleneck effect. Focusing on the models' perceptual capabilities, we observed that training on perception-related datasets with a lightweight VLM framework and integrating with Prism yielded remarkable performance, with the parameter count being only 1.8B.

% 借助Prism，我们在强依赖视觉的benchmark上评估了大量开源模型与商用API模型的内在感知能力，发现感知能力确实在限制着模型的能力，并且具有瓶颈效应。专注于模型的感知能力，我们发现使用轻量的VLM框架在感知相关的数据集上训练并接入Prsim可以获得很卓越的性能，并且它的参数量只有1.8B。

% 我们的工作以以下三点贡献为特征：1. 提出了一套显示的decouple框架用于解耦模型的感知推理，它本身高度灵活，可以调用不同的VLMs与LLMs完成image-text任务 2. 展现了全新的现有VLM的感知能力景观 3. 我们发现在prism框架下接入一个轻量级的感知模型可以获得一个remarkable的效果

In summary, the contributions of this work are as follows:
\begin{enumerate}[leftmargin=*,topsep=0.5ex,itemsep=-0.5ex,partopsep=0.75ex,parsep=0.75ex,partopsep=0pt,wide,labelindent=0pt]
    \item We introduce \textbf{Prism}, a highly adaptable framework designed to explicitly disentangle the perception and reasoning processes. 
    Prism enables the breakdown analysis of VLM capabilities and serves as a solution for vision-language tasks by integrating any given VLM and LLM.
    \item Utilizing Prism, we conduct a \textbf{decoupled analysis} of the perception and reasoning capabilities of existing VLMs. 
    Several intriguing findings emerge from the ablation study.
    \item Drawing inspiration from these findings, we integrate a lightweight VLM focused on perception with a powerful LLM dedicated to reasoning within the Prism framework. 
    Quantitative results demonstrate that this combination exhibits \textbf{outstanding performance and efficiency} across a range of vision-language tasks.
\end{enumerate}
\section{Methodology}
\subsection{Prism Architecture}
% solid on theory，lower bound & upper bound on specific backend to approximate the perception faculty of LVLMs

% Prism 框架以感知前端与推理后端模块化的结构为特征，将解答问题的过程拆解为两阶段，并且模块可以灵活更换，如图所示。在感知前端，LVLM将跟随指令生成关于图片输入的描述。在推理后端，文本based的LLM作为推理专家，将基于描述文本进行语言层面推理得出答案。sec.1展示了如何通过PRISM框架对VLM的感知能力进行评估。sec.2 details了注重感知能力的summarizer的训练以及到Prism的接入

Prism is characterized by a modular design that decomposes the process of solving visual questions into two stages, 
consisting of a perception module and a reasoning module, 
both of which can be flexibly replaced, as depicted in \cref{fig:Prism}. 
The perception module, typically a VLM,
initially follows the instruction to extract visual information from images and articulates this information in textual form.
The instruction can be generic or query specific, \emph{ie.}, written by the reasoning module given the question (text-only) as contextual information. 
Meanwhile, the reasoning module, usually an LLM, 
performs text-based reasoning on the textual information to generate answers to the questions. 
To assess the perception capabilities of various VLMs, 
we carefully select an appropriate benchmark corresponding to the principle we illustrate in \cref{sec:select-benchmark}. 
In \cref{sec:assessment},
we describe how we utilize the Prism framework to assess the perception and reasoning capabilities of VLMs, respectively. 
To substantiate our belief in the potential of combining a small-scale VLM with a powerful LLM, 
we train an approximately 2B-parameter VLM based on the LLaVA architecture to serve as a visual captioner and integrate it into Prism as the perceptual module, 
as detailed in \cref{sec:solver}.

\subsection{To Analyze with a Suitable Benchmark}
\label{sec:select-benchmark}

Numerous multimodal benchmarks \cite{liu2023mmbench,Fu2023MMEAC,li2023seed,lu2023mathvista,yue2023mmmu,chen2024we} exist,
each evaluating the capabilities of VLMs from various perspectives. 
To maximize the utility of the Prism framework,
careful selection of the benchmark is imperative for the decoupling analysis.
In summary, we adhere to the following principles for benchmark selection: 
1) \textbf{Vision Indispensability}: The benchmark must require visual information for question solving, and questions that can be answered without utilizing visual information are excluded;
2) \textbf{Minimal Data Leakage}: The visual questions should not be part of the model's training data;
and 3) \textbf{Complexity}: The solving process of the visual question should involve both perception and reasoning components. 
Considering these three principals, 
we select MMStar~\cite{chen2024we} as the primary benchmark for the decoupling analysis and ablation. 
MMStar ensures vision indispensability and makes a concerted effort to minimize data leakage, while many other benchmarks are plagued by the two issues.

\subsection{Prism as an Analytical Framework}
\label{sec:assessment}

Prism functions as an analytical framework for evaluating the perception and reasoning capabilities of a given VLM. 
In this section, we elaborate on the methodology for evaluating these capabilities.

To assess the perception capabilities of various VLMs, 
we employ ChatGPT (GPT-3.5-turbo-0125) as the reasoning module and standardize instructions for description within Prism. 
With the reasoning module and instructions fixed, 
the final accuracy of the visual questions is solely determined by the quality of the visual information extracted. 
Under the controlled setting, we consider the VQA accuracy as a proxy to measure the perception capability of VLMs. 

Within the Prism framework, the instructions for visual information extraction
are crucial as they are designed to elicit the fundamental perceptual capabilities of VLMs.
We have adopted two types of instructions to assess the perceptual abilities of models:
\begin{enumerate}
[leftmargin=*,topsep=0.5ex,itemsep=-0.5ex,partopsep=0.75ex,parsep=0.75ex,partopsep=0pt,wide,labelindent=0pt]
    \item \textbf{Generic Instruction}: a standardized, universal instruction aimed at extracting and describing the basic elements present in an image;
    \item \textbf{Query-Specific Instruction}: a combination of the generic instruction and an incremental instruction that directs the VLM to provide a detailed account of the visual information relevant to the question. 
    The incremental instruction is crafted by the reasoning module given the text-only question.
\end{enumerate}

Employed with the generic instruction, 
Prism offers a straightforward decoupling pipeline. 
Meanwhile, using query-specific instructions is a more realistic setting and can better realize the full potential of the Prism framework. 
A Prism decoupled GPT-4o (GPT-4o for both perception and reasoning) 
achieves almost the same quantitative performance compared to the end-to-end GPT-4o. 
Analyzing the results, we find that with the reasoning module and instruction controlled,
VLMs with language models of different sizes display a much narrowed gap in perception performance compared to the results under the end-to-end setting.

Besides evaluating the perception performance, 
Prism can also roughly measure the reasoning capabilities of VLMs.
When used to solve general visual questions in an end-to-end manner, 
the VLM implicitly performs reasoning with its language encoder. 
Alternatively, Prism provides an explicit pipeline, 
in which an off-the-shelf LLM is separately used to predict the answer based on the visual information. 
For small-scale VLMs (7B-parameter, \textit{etc.}), 
we find that Prism equipped with the VLM and ChatGPT-3.5 outperforms the end-to-end VLM in quantitative performance, especially for reasoning related VQA.
The results reveal that for small-scale VLMs, the overall performance can be heavily constrained by the parameter size of the language model. 

% Confirming perception contents of the question relies on GPT-4's few-shot\footnote{GPT-4 version: gpt-4-0125-preview}, text-only responses based on the question. Compared to using GPT-4V to confirm relevant contents based on text-image pairs, this approach avoids the inclusion of visual detail information that might be present in the responses.
% 关于推理后端，我们发现不同的LLM存在不同的pattern。总的来说，一些严格而且保守的模型更倾向于在解决text-based reaoning任务时认为描述信息不够而拒答。所以我们在inference prompt中添加指令，使LLM尽量选择一个最贴切的选项。对于剩下的拒答的情况，我们采取random choice为backup以减小这种pattern带来的bias
% Regarding the reasoning backend, we find that different LLMs exhibit distinct patterns. Generally, some strict and conservative models tend to refuse to answer text-based reasoning tasks when they consider the descriptive information insufficient. Therefore, we have added instructions to the inference prompt to encourage LLMs to choose the most appropriate option available. For the remaining instances where the model refuses to answer, we adopt a random choice as a backup to minimize the bias introduced by this pattern.

\subsection{Prism as a Vision-Language Task Solver}

\label{sec:solver}
Beyond serving as an evaluation framework, 
Prism can also function as an \textbf{efficient vision-language task solver}.
The perception module in Prism can incorporate one or multiple VLMs
to extract high-quality visual information. 
Concurrently, the reasoning module can be instantiated with a powerful LLM to harness its advanced reasoning capabilities. 

% Meanwhile, Prism can contain more than one VLM in perception module and combine the visual information of different VLMs, which can obtain progressive performance, compared to using a single VLM.

In \cref{sec:assessment}, 
we observed that VLMs paired with language models of varying sizes exhibit similar performance in perception capability. 
This suggests that it is promising to employ small-scale VLMs to generate informative visual descriptions, 
serving as an efficient perception module. 
To validate this concept, 
we conduct extensive experiments to train visual captioners, 
utilizing the widely adopted LLaVA architecture and only open-source datasets. 
Below, we elaborate on the specific settings considered for this ablation study in detail:

% Addtionally, Prism ensures the feasibility of training VLMs as visual captioners to solve vision-language tasks when LLM is easily accessible today. We adopt the vanilla LLaVA architecture and conduct ablation and training of Prism Captioners with the following strategies:

\textbf{Instruction Tuning Data.} 
We use \texttt{ALLaVA-Caption-4V} and \texttt{Evol-Intruct-GPT4-Turbo-143K} in \texttt{ALLaVA}~\cite{chen2024allava} as our instruction tuning data. 
The former comprises 715K pairs of images and detailed visual captions produced by GPT-4v, 
whereas the latter contains 143K text instruction tuning data, generated by GPT-4-Turbo.
Utilizing the descriptive data for instruction tuning better triggers the VLM's ability to extract and articulate visual information more effectively, 
compared to instruction tuning data in QA formats \cite{liu2023visual,Liu2022VisualSR,textvqa,masry2022chartqa}. 

\textbf{Model Architecture. } 
To investigate the impact of vision encoder in LLaVA, 
we experimented with multiple encoders, including \texttt{CLIP ViT-L/14}~\cite{radford2021learning}, \texttt{SigLip-SO400M}~\cite{zhai2023sigmoid}, and \texttt{InternViT-6B}~\cite{chen2023internvl}. 
For the language encoder, 
we tested two lightweight variants of InternLM2~\cite{cai2024internlm2}: 
\texttt{InternLM2-7B} and \texttt{InternLM2-1.8B}. 
The inference of all combinations (excluding \texttt{InternViT-6B}) can be efficiently executed on consumer-level GPUs such as RTX 4090, \emph{etc.} 
During instruction tuning, we maintain the vision encoder fixed 
and apply QLoRA~\cite{dettmers2023qlora} to the language encoder.

% \textbf{Training.} Considering efficiency, one-stage training is easier to implement and less time-consuming than multi-stage training, even though the latter may be more popular~\cite{liu2024llavanext, liu2023visual, chen2023internvl, chen2024far}. Meanwhile, we perform QLoRA, training the projector module and low-rank adapters with LLM quantized and vision encoder freezed, instead of fully finetuning~\cite{he2024efficient}.

Regarding the vision backbone, \texttt{SigLip} exhibits relatively superior performance. 
Additionally, we observed that a larger language model only results in minor differences in perception performance.
Our findings indicate that, within the Prism framework, a 2B vision captioner 
can achieve strong perception performance on par with LLaVA-NeXT~\cite{liu2024llavanext} equipped with a 34B language backbone.

\section{Evaluation Results}
\label{sec:experiments}

\subsection{Implementation Details}
\label{sec:eval-settings}
\textbf{Evalution Details.} 
We use Prism to evaluate the capabilities of various VLMs, 
which can be categorized into two major groups:
(a) \textbf{Proprietary VLMs}, including GPT-4o (20240513)~\cite{OpenAI2023GPT4TR}, GPT-4v (20231106)~\cite{OpenAI2023GPT4TR}, GeminiPro-V~\cite{team2023gemini}, and Qwen-VL-Max~\cite{bai2023qwen}; 
(b) \textbf{Open-Source VLMs}, including LLaVA-v1.5~\cite{liu2023improved}, InternLM-XComposer2~\cite{internlmxcomposer2}, mPLUG-Owl2~\cite{ye2023mplugowl2}, LLaVA-NeXT~\cite{liu2024llavanext}, InternVL-Chat-v1.5~\cite{chen2024far}, DeepSeek-VL~\cite{lu2024deepseek}, MiniCPM-V-2~\cite{minicpm2024}. 
When integrating these VLMs as perception modules within Prism,
we employ greedy decoding and limit the maximum number of output tokens to 512. 
The evaluation encompasses both generic and query-specific instructions.
Unless otherwise specified, GPT-3.5-Turbo-0125 is adopted as the reasoning module. 
All evaluations are conducted using VLMEvalKit~\cite{2023opencompass}. 
Further details on the evaluation are provided in \cref{sec:evalution-details}. 

\textbf{Post-Processing. } 
Within the Prism framework, the LLM (particularly proprietary APIs) in the reasoning module
often declines to answer questions due to insufficient clues in the extracted visual information. 
When Prism is employed as an evaluation framework,
we classify this as a failure of perception for the visual question and refrain from any post-processing. 
For fair comparisons,
whether Prism is contrasted with other VLMs as a vision-language task solver or
different LLMs are evaluated as reasoning modules with varying rejection rates, 
a random choice is utilized as a fallback when option matching fails.

\begin{table*}[t]
\vspace{-5mm}
\centering
% \vskip -1.1em
% Result of Instruction\emph{Query-Specific Instruction} and \emph{Generic Instruction}
\resizebox{1.0\linewidth}{!}{
\tablestyle{5pt}{1.4}
\begin{tabular}{l|cccccc|c|cccccc|c}
\toprule
\multirow{2.5}{*}{\bf Model} & \multicolumn{7}{c|}{{\cellcolor{maroon}}{\bf Generic Instruction}} &  \multicolumn{7}{c}{{\cellcolor{lightblue}}{\bf Query-Specific Instruction}}\\
\cmidrule(lr){2-8} \cmidrule(lr){9-15} 
& \textbf {CP} & \textbf {FP} & \textbf {IR} & \textbf {LR} & \textbf {Math} & \textbf {ST} & \textbf {Overall} & \textbf {CP} & \textbf {FP} & \textbf {IR} & \textbf {LR} & \textbf {Math} & \textbf {ST} & \textbf {Overall} \\

\midrule
\rowcolor{gray!15}\multicolumn{15}{c}{\emph{Proprietary VLMs}}  \\
\bf GPT-4o~\cite{OpenAI2023GPT4TR} & \underline{64.0} & \bf 41.6 & \underline{54.4} & \bf 51.6 & \bf 47.6 & 31.2 & \bf 48.4 & \bf 67.2 & \bf 51.6 & \bf 63.2 & \bf 56.4 & \bf 52.0 & \bf 36.8 & \bf 54.5 \\
\bf GPT-4v~\cite{OpenAI2023GPT4TR} & 62.4 & 33.2 & 51.2 & 39.2 & 44.4 & \bf 32.8 & 43.9 & \bf 67.2 & 43.6 & 56.8 & 46.8 & 42.0 & 30.8 & 47.9 \\
\bf GeminiPro-V~\cite{team2023gemini} & 57.2 & 36.0 & 51.6 & 43.2 & 45.6 & 23.2 & 42.8 & 61.2 & 35.2 & 53.2 & 47.6 & 48.4 & 19.2 & 44.1 \\
\bf Qwen-VL-Max~\cite{qwen} & 62.4 & 28.8 & 49.2 & 45.2 & 37.2 & 21.2 & 40.7 & 62.4 & 36.0 & 54.8 & 46.4 & 38.0 & 24.0 & 43.6 \\
\rowcolor{t_green}\multicolumn{15}{c}{\emph{Open-Source VLMs}} \\
\bf InternVL-Chat-v1.5~\cite{chen2024far} & 59.2 & 33.2 & 51.2 & \underline{42.8} & 40.4 & \underline{30.4} & \underline{42.9} & \underline{65.6} & \underline{42.4} & 56.4 & \underline{50.8} & \underline{46.8} & \underline{28.0} & \underline{48.3} \\
\bf InternLM-XComposer2~\cite{internlmxcomposer2} & 56.0 & 32.0 & \bf 54.8 & 33.6 & 37.2 & 20.8 & 39.1 & 60.8 & 39.2 & \underline{58.4} & 47.6 & 42.4 & 19.6 & 44.7 \\
\bf LLaVA-NeXT (Yi-34B)~\cite{liu2024llavanext} & \bf 64.4 & \underline{37.2} & 50.0 & 34.0 & 39.2 & 24.8 & 41.6 & 60.0 & 41.6 & 52.4 & 44.0 & 40.8 & 25.2 & 44.0 \\
\bf DeepSeek-VL-7B~\cite{lu2024deepseek} & 57.2 & 31.6 & 49.6 & 42.0 & \underline{43.6} & 24.0 & 41.3 & 61.2 & 34.8 & 54.4 & 42.8 & 45.2 & 25.2 & 43.9 \\
\bf LLaVA-NeXT (Mistral-7B)~\cite{liu2024llavanext} & 60.8 & 30.8 & 49.6 & 37.6 & 36.8 & 20.4 & 39.3 & 62.8 & 36.4 & 53.6 & 42.0 & 35.2 & 26.4 & 42.7 \\
\bf MiniCPM-V-2~\cite{minicpm2024} & 55.6 & 28.4 & 49.2 & 39.6 & 38.4 & 19.2 & 38.4 & 61.2 & 30.4 & 52.0 & 44.4 & 40.8 & 22.8 & 41.9 \\
\bf LLaVA-NeXT (Vicuna-13B)~\cite{liu2024llavanext} & 60.8 & 38.0 & 48.8 & 35.2 & \underline{43.6} & 20.8 & 41.2 & 63.2 & 38.8 & 49.2 & 38.8 & 37.6 & 23.2 & 41.8 \\
\bf LLaVA-NeXT (Vicuna-7B)~\cite{liu2024llavanext} & 63.2 & 28.8 & 46.4 & 40.8 & 34.0 & 22.8 & 39.3 & 56.8 & 37.6 & 47.2 & 42.4 & 36.4 & 21.6 & 40.3\\
\bf mPLUG-Owl2~\cite{ye2023mplugowl2} & 47.2 & 24.0 & 45.2 & 33.6 & 28.8 & 18.8 & 32.9 & 53.2 & 34.4 & 45.2 & 38.8 & 38.8 & 23.6 & 39.0 \\
\bf LLaVA-v1.5-13B~\cite{liu2023improved} & 45.2 & 28.4 & 45.2 & 35.6 & 30.0 & 17.2 & 33.6 & 54.8 & 28.8 & 48.8 & 37.6 & 32.0 & 20.0 & 37.0 \\
\bf LLaVA-v1.5-7B~\cite{liu2023improved} & 48.8 & 29.6 & 48.0 & 34.4 & 27.6 & 17.2 & 34.3 & 50.8 & 31.6 & 50.4 & 38.8 & 32.0 & 25.2 & 38.1 \\
% \bf IDEFICS2-8B\cite{laurençon2024matters} & 47.2 & 26.0 & 42.0 & 40.8 & 37.6 & 20.8 & 35.7 & 49.6 & 27.2 & 47.2 & 37.6 & 43.2 & 21.2 & 37.7 \\
\bottomrule
\end{tabular}
}
\vspace{-1mm}
\caption{\textbf{Detailed Perception Performance on MMStar under Prism Evaluation Framework}. Reasoning module: ChatGPT. Abbreviations: Coarse Perception (CP), Fine-grained Perception (FP); Instance Reasoning (IR); Logical Reasoning (LR); Science\&Technology (ST). 
Overall best scores are marked as \textbf{bold}, and intra-category best scores are marked as
\underline{underline}. }
% Models are sorted by the descending order of overall score with query-specific instruction. 
\label{tab:main-result}
\vspace{-5mm}
% \vskip -0.5em
\end{table*}

% \begin{table*}[h]
% \centering
% \label{tab: performance}
% \resizebox{1.0\linewidth}{!}{
% \tablestyle{5pt}{1.4}
% \begin{tabular}
% \end{tabular}
% }
% \caption{Result of Perception Model-Inference Model }
% \label{tab:main-result}
% % \vskip -0.5em
% \end{table*}

\subsection{Main Results}

We present the primary evaluation results of various VLMs' perception performance in \cref{tab:main-result}.  
Results are categorized under two instruction types: \textit{\colorbox{maroon}{Generic Instruction}} and \textit{\colorbox{lightblue}{Query-Specific Instruction}}.

\textbf{Generic Instruction Results. } 
When using generic instructions, 
GPT-4o exhibits exceptional performance across a variety of tasks. 
InternVL-Chat-v1.5 achieves the highest overall perception performance among open-source VLMs, 
nearly on par with the proprietary GPT-4v. 
LLaVA-NeXT (Yi-34B) demonstrates strong coarse and fine-grained perceptual abilities but lags behind GPT-4v in recognizing abstract elements in LR and ST visual questions. 
Smaller open-source VLMs, such as mPLUG-Owl2 and LLaVA-v1.5-7B, 
struggle with both coarse and fine-grained perception and 
encounter difficulties in recognizing essential elements in reasoning-related VQA.

\textbf{Query-Specific Instruction Results.} 
GPT-4o demonstrates significantly superior performance compared to all other models in the extraction and expression of visual information across all dimensions. 
GPT-4v is equally adept at handling coarse perception contents. 
Among open-source VLMs, InternVL-Chat-v1.5 excels in perception tasks across all dimensions excluding LR. 
Its performance not only surpasses that of other open-source VLMs but is also slightly ahead of GPT-4v. 
Occasionally, a decline in performance can be observed for specific VLM-task pairs when using query-specific instructions instead of generic ones, 
which can stem from difficulty in understanding query-specific instructions for specific domains.

\subsection{Detailed Analysis}

\textbf{Are Proprietary Models Getting Ahead of the Game?}
When employing both generic and query-specific instructions, proprietary VLMs, 
particularly GPT-4o, significantly surpass other models in perceptual capabilities and 
can adeptly manage a wide range of tasks, as demonstrated in \cref{tab:main-result}.
Certain open-source models, such as InternVL-Chat-v1.5 and LLaVA-NeXT (Yi-34B), have achieved notable performance, approaching the capabilities of proprietary VLMs like GPT-4v and GeminiPro-V.
Other open-source models, due to their limited perceptual abilities, 
generally perform slightly worse in Math and ST assessments. 
Notably, MiniCPM-V-2, a lightweight VLM with $\sim$3B parameters, 
displays better perceptual performance compared to some 7B VLMs.

\begin{figure}[t]
  \vspace{-5mm}
  \centering
  \includegraphics[width=\textwidth]{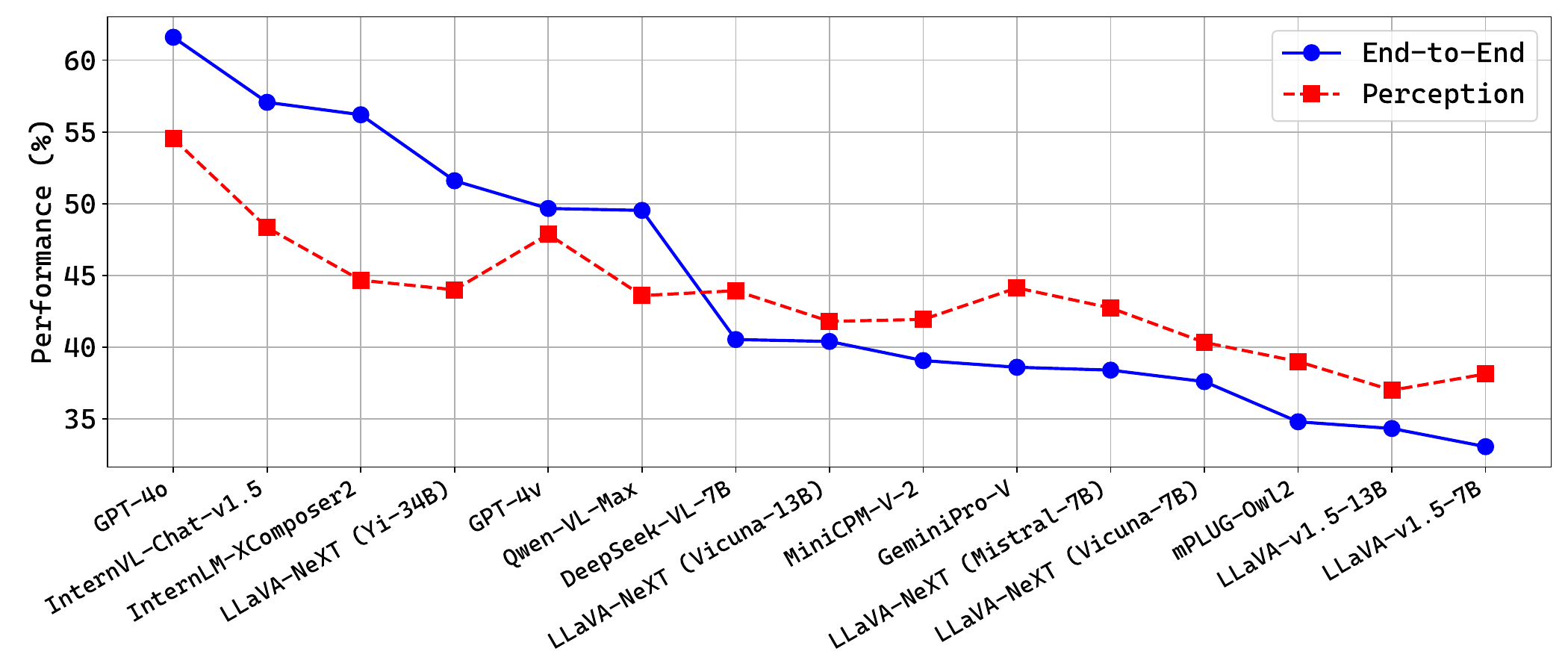}
  \vspace{-6mm}
  \caption{\textbf{Comparing End-to-End Performance and Perception Capability on MMStar.} We display model accuracies in end-to-end VQA and the Prism perception test with query-specific instructions. 
  Most small-scale (7B, 13B, \textit{etc.}) VLMs achieve better performance within Prism. }
  \label{fig:e2evsprism}
  \vspace{-6mm}
\end{figure}

\textbf{The Gap between Perception Ability and End-to-End Performance. }
In addition to solving visual questions in an end-to-end manner, 
Prism provides an alternative pipeline where the VLM is solely utilized for perception.
The distinction between these two methods lies in the reasoning process: 
the former conducts reasoning internally within VLMs,
whereas the latter performs reasoning based on VLM extracted information using an external LLM (ChatGPT). 
The comparison between these two approaches on MMStar is depicted in \cref{fig:e2evsprism}.
For state-of-the-art large-scale VLMs such as GPT-4o and InternVL-Chat-v1.5, 
which are expected to possess excellent reasoning capabilities, 
employing an external ChatGPT for reasoning may diminish overall performance.
Conversely, for most small-scale VLMs,
using ChatGPT for reasoning significantly improves their performance,
particularly in reasoning-related VQA, as shown in \cref{fig:reasoning_gap}. 
This phenomenon indicates that the overall performance of small-scale VLMs can be heavily constrained by the size of the language model. 
To investigate whether the reasoning ability of ChatGPT constrains state-of-the-art VLMs, 
we implemented a Prism pipeline that decouples GPT-4o by using it as both the perception and reasoning module.
The result reveals that this Prism pipeline, with post-processing, 
achieves an overall accuracy of 61\%, 
nearly identical to the end-to-end GPT-4o performance of 61.6\%.

\begin{figure}[t]
  \centering
  \includegraphics[width=\textwidth]{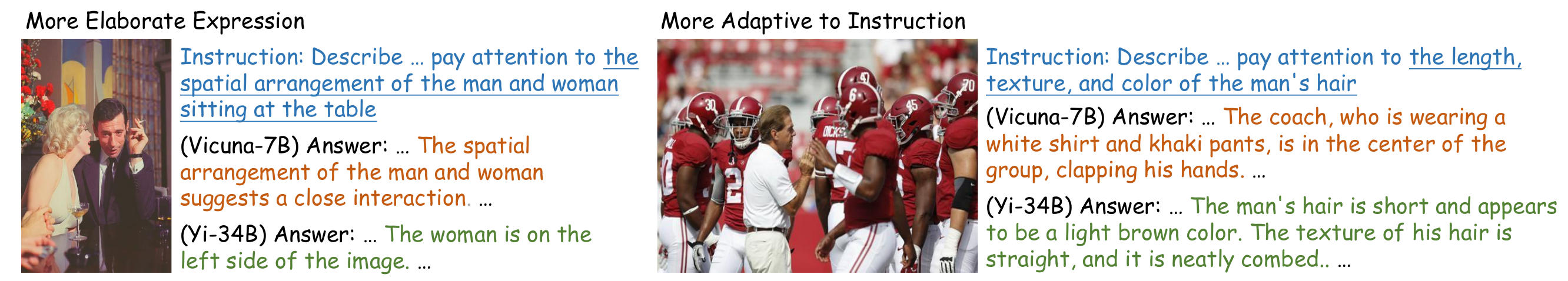}
  \vspace{-5mm}
  \caption{\textbf{The Effect of Language Model Size on Perception Ability.} 
  We compare visual information extracted from different LLaVA-NeXT models. 
  \textbf{Left}: LLaVA-NeXT (Yi-34B) tells the spatial arrangement in a more detailed way;
  \textbf{Right}: LLaVA-NeXT (Vicuna-7B) dismisses the query on the man's hair while LLaVA-NeXT (Yi-34B) tells all contents elaborately following the instruction.}
  \label{fig:LLM_SIZE_CASE}
  \vspace{-2mm}
\end{figure}

\textbf{How does Language Model Size Affect Perception Ability? }
During evaluation, 
we observe that the LLaVA-v1.5 series shows no significant improvement when using a larger language model (Vicuna-13B instead of Vicuna-7B, \textit{etc.}). 
This suggests that perception performance may be independent of the language model size when using a relatively low-resolution vision backbone. 
However, the situation appears to differ with LLaVA-NeXT. 
Quantitative results for the LLaVA-NeXT series tell that scaling up the language model slightly enhances model perception, 
particularly when using query-specify instructions. 
Through a detailed qualitative analysis, 
we identified the primary factors contributing to the superior performance of larger LLaVA-NeXT models over smaller ones as follows:
(1) \textbf{More Elaborate Expression}: Models equipped with a larger language encoder exhibit enhanced ability to articulate visual information. 
More detailed and organized narratives make it easier for the reasoning module to answer the question;
(2) \textbf{More Adaptive to Instruction}:  
Larger language backbones entitle the model with a better understanding of instructions, 
yielding more suitable textual visual information for reasoning, 
particularly in response to query-specific instructions. 
In \cref{fig:LLM_SIZE_CASE}, we provide some qualitative results about the two typical modes.

\section{Ablation Study}

\begin{table}[t]
    \vspace{1mm}
    \begin{minipage}{.36\linewidth}
        \centering
        \resizebox{\linewidth}{!}{
        \tablestyle{5pt}{1.2}
        \begin{tabular}{ccc}
        \toprule
        \bf Instructions & \bf GPT-4o & \bf \makecell{LLaVA-NeXT\\(Yi-34B)} \\
        \midrule
        \bf Human 2 & 47.7 & 40.4 \\
        \bf GPT Synthesize 1 & 47.1 & 41.3 \\
        \bf GPT Synthesize 2 & 48.1 & 40.1 \\
        \bf CoT & 47.7 & 41.4 \\
        \bf Decompose & 47.3 & 41.4 \\
        \midrule
        \bf Human 1 & \bf 48.4 & \bf 41.6 \\
        \bottomrule
        \end{tabular}
        }
        \vspace{2mm}
        \caption{\textbf{Ablation on Different Generic Instructions. }}
        \label{tab:prompt_detail}
    \end{minipage}%
    \hfill
    \begin{minipage}{.60\linewidth}
      \centering
        \resizebox{\linewidth}{!}{
        \tablestyle{5pt}{1.5}
        \begin{tabular}{ccccc}
        \toprule
        \bf Model & \bf GPT-4o & \bf GPT-4v & \bf \makecell{LLaVA-NeXT\\(Yi-34B)} & \bf LLaVA-v1.5-7B \\
        \midrule
        \bf GPT-3.5-Turbo-0125  & 54.7 & 48.5 & 44.1 & \textbf{38.5}   \\
        \bf GPT-4-Turbo-0125 & 56.9 & 48.7 & \textbf{48.4} & 38.3\\
        \bf Llama-3-70B-Instruct & \textbf{59.3} & \textbf{51.3} & 47.7 & \textbf{38.5} \\
        \bf DeepSeek-v2-Chat & 57.8 & 49.7 & 45.8 & 35 \\
        \bottomrule
        \end{tabular}
        }
        \vspace{3mm}
        \caption{\textbf{Ablation on Using Different LLMs as the Reasoning Module.}}
        \label{tab:reason_ablation}
    \end{minipage} 
    \vspace{-6mm}
\end{table}

\textbf{Ablation on Generic Instructions. } 
Within Prism,
the generic instructions for visual information extraction are crucial. 
We experimented with a variety of instructions to elicit the fundamental perceptual capabilities of VLMs,
including human-written instructions, GPT-4 generated instructions, and those incorporating chain-of-thought~\cite{wei2023chainofthought} or explicit decomposition,
as shown in \cref{fig:select-instruction}. 
We conducted an ablation study on MMStar using the state-of-the-art VLM GPT-4o  and LLaVA-NeXT (Yi-34B). 
As illustrated in \cref{tab:prompt_detail}, 
\texttt{Human 1} outperforms others in eliciting the fundamental perceptual capabilities of the models, 
while the differences among various instructions are not significant. 
Therefore, we adopt \texttt{Human 1} as the generic instruction for all evaluations. 

\begin{figure*}[t] 
% \vspace{-5mm}
\begin{AIbox}{Different Generic Instructions}
\fontsize{8pt}{10pt}\selectfont
{
     \textcolor{red}{\textbf{Human 1:}} Describe the fine-grained content of the image, including scenes, objects, relationships, instance location, and any text present.
    
    \textcolor{blue}{\textbf{Human 2:}} Describe the fine-grained content of the image, including scenes, objects, relationships, instance location, background and any text present. Please skip generating statements for non-existent contents and describe all you see.
    
    \textcolor{blue}{\textbf{GPT Synthesize 1:}} Given the image below, please provide a detailed description of what you see.

    \textcolor{blue}{\textbf{GPT Synthesize 2:}} Analyze the image below and describe the main elements and their relationship.

    \textcolor{blue}{\textbf{CoT:}} Describe the fine-grained content of the image, including scenes, objects, relationships, instance location, and any text present. Let's think step by step.

    \textcolor{blue}{\textbf{Decompose:}} Decompose the image into several parts and describe the fine-grained content of the image part by part, including scenes, objects, relationships, instance location, and any text present.
    
}
\end{AIbox} 
\vspace{-2mm}
\caption{\textbf{Different Generic Instructions we adopted in the Ablation Study. }}
\label{fig:select-instruction}
\vspace{-2mm}
\end{figure*}

\textbf{Ablation on the Reasoning Module. }
The reasoning module is critical for accurately determining the correct answer based on the visual information.
To evaluate the impact of the reasoning module on overall performance,
we select four LLMs:
\texttt{GPT-3.5-Turbo-0125}, \texttt{GPT-4-Turbo-0125}, \texttt{Llama-3-70B-Instruct}, and \texttt{DeepSeek-v2-Chat}, 
and assess these models across four VLMs with varying capacities. 
The comparative results are presented in \cref{tab:reason_ablation}. 
As a freely available model, GPT-3.5 demonstrates good reasoning performance under our framework, 
and the more advanced GPT-4 shows improved performance in line with other benchmarks. 
Notably, \texttt{Llama3-70B-Instruct}, 
as a representative of open-source models, 
exhibits competitive capabilities compared to \texttt{GPT-4-Turbo-0125} under different perceptual conditions. 
\noindent
\begin{table}[t]
\begin{minipage}{0.7\linewidth}
\centering
\includegraphics[width=\textwidth]{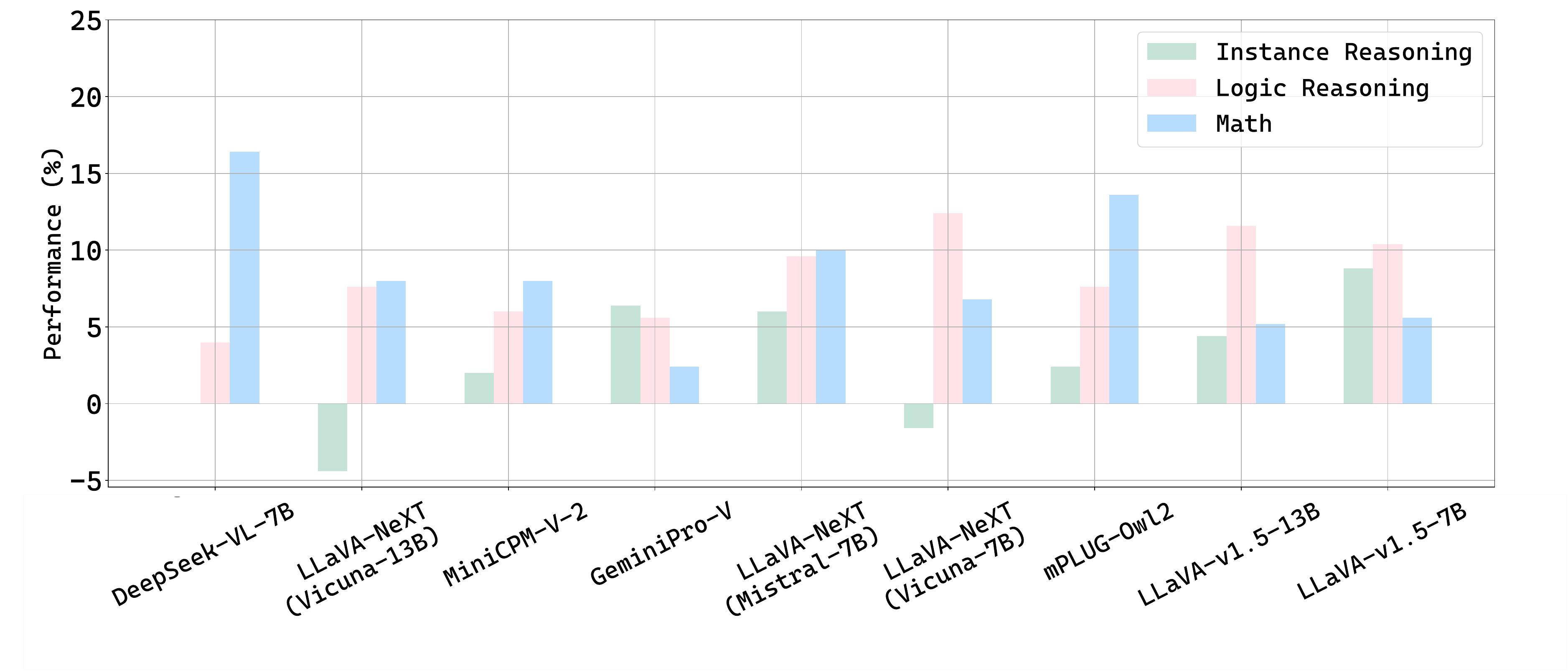}
\vspace{0.02mm}
\captionof{figure}{\textbf{The Performance Changes of Using an External LLM (ChatGPT) for Reasoning of Small Scale VLMs. }}
\label{fig:reasoning_gap}
\end{minipage}
\hfill
\begin{minipage}{0.28\linewidth}
\centering
\vspace{10mm}
\resizebox{\linewidth}{!}{
\tablestyle{10pt}{1.5}
\begin{tabular}{cc}
\toprule
\bf \makecell{Vision\\Backbone} & \bf Overall \\
\midrule
\bf SigLip-SO400M & \textbf{43.5} \\
\bf InternViT-6B & 42.7 \\
\bf CLIP ViT-L/14 & 41.7 \\
\bottomrule
\end{tabular}}
\vspace{12mm}
\captionof{table}{\textbf{Ablation on the Vision Backbone.}}
\label{tab:vision_ablation}
\end{minipage}
\vspace{-5mm}
\end{table}
This suggests that open-source models could be valuable for further exploration in visual reasoning.

% which may signal that open-source models can be considered to be utilized for further explorations of visual reasoning, while DeepSeek-Chat also demonstrates a noteworthy level of performance and is more easily invoked through API.
% To compare the effect of the reasoning module on the final performance, we choose four LLMs (including ChatGPT, GPT-4-0125, Llama3-70B-chat, and DeepSeek-Chat), under four LVLMs with differential capacity, to compare their ability, and the results are shown in the table. 
% reasoning部分也在prism框架下有重要作用，为了比较推理模型对最终表现的影响，我们选用了四个大语言模型（包括..），在四个不同能力的VLM下，来比较他们的推理表现，结果如表所示。 chatgpt是free available的模型，在我们这个框架评估下展现出了不错的性能，而更新一代的GPT-4则表现出了与其时间相匹配的更强的性能。令人惊喜的是，llama3-70b-chat作为开源模型的代码，在不同感知情况下均能展现出与GPT-4有强竞争力的推理性能，这也许预示着可以考虑利用开源模型来进行视觉推理的进一步探索，同时Deepseek-chat也展现了不俗的水平，且更易通过API进行调用。
% \input{tables/detailed}
\textbf{Ablation on the Vision Backbone. }
% The extraction of visual features by the vision backbone has a significant impact on the perception module. Hence, we utilized two powerful vision backbones, SigLip and InternVIT, and followed the same configuration for the ablation experiments. The results, shown in \Cref{tab:vision_ablation}, demonstrate that SigLip has a more noticeable performance improvement compared to the CLIP and InternVIT, which may be since it employs a loss function that matches the visual and text at a finer granularity and thus can generate more apt descriptions.
To investigate the impact of the vision encoder on perception ability within the LLaVA architecture, 
we conduct an ablation study on three pre-trained visual backbones, 
including \texttt{CLIP ViT-L/14}, \texttt{SigLip-SO400M}, and \texttt{InternViT-6B}. 
We use \texttt{InternLM2-7B} as the fixed language encoder in LLaVA and train the VLMs for one epoch with the vision encoder fixed.  
The results in \cref{tab:vision_ablation} show that \texttt{SigLip-SO400M} achieves better performance compared to \texttt{CLIP ViT-L/14} and \texttt{InternViT-6B} on MMStar.

% This improvement may be due to SigLip's use of a loss function that aligns visual and text features at a finer granularity, enabling it to generate more accurate descriptions.

 % Therefore, we used two powerful vision backbones, SigLip and InternVIT, and maintained the same configuration for our ablation experiments.

\section{PrismCaptioner}

Within Prism, 
we explore the use of small-scale Vision-Language Models (VLMs) 
as a perception module.
We use \texttt{SigLip} as the vision encoder, 
\texttt{InternLM2-[1.8B/7B]} as the language encoder to 
develop two visual captioners at different scales, 
refered as \texttt{PrismCaptioner-[2B/7B]}.

\subsection{Training Details}
\label{sec:training-details}

We perform one-stage training for two epochs using ZeRO2 with XTuner~\cite{2023xtuner} on 8 NVIDIA A800-80GB GPUs, and the training lasts less than a day. 
The training data include of one copy of \texttt{ALLaVA-Caption-4V} and two copies of \texttt{Evol-Intruct-GPT4-Turbo-143K}.
The batch size is set to 128 for \texttt{PrismCaptioner-2B} and 64 for \texttt{PrismCaptioner-7B}. 
We utilized the AdamW optimizer and 2e-4 learning rate, 
with the warm-up ratio set to 0.03 and ($\beta_1, \beta_2$) set to  (0.9, 0.999). 
No weight decay was applied and a maximum norm value of 1 is applied for gradient clipping. Full details about QLoRA are presented in \cref{sec:lora-details}.
% These settings will cost 78GB CUDA memory at peak. In terms of time-consuming, \texttt{SigLip} brings more time-cost burdens that \texttt{CLIP}. Training PrismCaptioner-2B for one epoch (994K) takes about 7 hours, while training PrismCaptioner-7B for one epoch (994K) costs about 14 hours.

\subsection{The Performance of PrismCaptioner}
\label{sec:captioner-performance}

We conduct a thorough evaluation of \texttt{PrismCaptioners} across multiple benchmarks, 
employing \texttt{GPT-3.5-Turbo-0125} and \texttt{Llama-3-70B-Instruct} as the reasoning module. 
We utilize MMStar as our primary benchmark to
assess comprehensive multimodal capabilities. 
For domain-specific evaluations,
we choose AI2D to gauge diagram comprehension, 
MMMU for expert knowledge assessment, and MathVista to test mathematical proficiency. 
The results are presented in \cref{tab:detail-performance-prism}. 
In line with the benchmark selection principles outlined in \cref{sec:select-benchmark}, 
we apply a consistent filtering strategy across AI2D, MMMU, and MathVista, 
mirroring that of MMStar, to retain only vision-essential and uncontaminated questions, 
denoted by the suffix (F). 
In addition to comparisons with existing open-source and proprietary VLMs, 
we also assess two baseline models, \texttt{LLaVA-InternLM2-[1.8B/7B]}, 
which are \texttt{PrismCaptioners} trained on LLaVA-v1.5 instruction tuning data. 
We also compare \texttt{PrismCaptioners} and \texttt{ShareCaptioner}, 
an open-source VLM designed for generating informative image captions, 
under identical Prism framework configurations.

As depicted in \cref{tab:detail-performance-prism},
each \texttt{PrismCaptioner},
with an external potent LLM for reasoning, 
markedly surpasses its corresponding end-to-end baseline.
\texttt{PrismCaptioners} also outperform \texttt{ShareCaptioner} across all multimodal benchmarks. 
For the 7B variant,
the integration of \texttt{Llama3} results in a substantial enhancement, 
positioning \texttt{PrismCaptioner-7B} as a highly competitive vision-language solver, 
particularly on MMStar and MMMU. 
For \texttt{PrismCaptioner-2B}, 
employing \texttt{ChatGPT} yields superior results, 
outperforming nearly all 7B VLMs in general aptitude, expert knowledge, and mathematical skills. 
Remarkably, it achieves performance levels on par with some ten times larger VLMs, such as \texttt{LLaVA-InternLM2-20B}, \texttt{Yi-VL-34B}, and \texttt{Emu2-Chat}. 
This demonstrates that Prism enables the creation of a robust yet efficient vision-language solver, 
exemplified by \texttt{PrismCaptioner-2B} with \texttt{ChatGPT}, which delivers impressive results.

\begin{table*}[t]
\centering
% \vskip -1.1em
% Result of Instruction\emph{Query-Specific Instruction} and \emph{Generic Instruction}
\resizebox{1.0\linewidth}{!}{
\tablestyle{7pt}{1.4}
\begin{tabular}{l|c|cc|cc|cc}
\toprule
\bf Model & \textbf{MMStar} & \textbf{MMMU} & \makecell{\textbf{MMMU (F)}} & \textbf{MathVista} & \makecell{\textbf{MathVista (F)}} & \textbf{AI2D} & \makecell{\textbf{AI2D (F)}} \\
\midrule
\rowcolor{gray!15}\multicolumn{8}{c}{\emph{Proprietary VLMs}}  \\
\bf GPT-4o~\cite{OpenAI2023GPT4TR} &\bf 61.6&	\bf 62.8	&\bf 45.9&	\underline{56.5}&	\underline{46.5}&	\bf 82.2	& \bf 59.7 \\
\bf GPT-4v~\cite{OpenAI2023GPT4TR} &49.7	&53.8	&42.0	&48.7&	32.0&	75.9&	45.7\\
\rowcolor{t_green}\multicolumn{8}{c}{\emph{Open-Source VLMs}} \\
\bf InternVL-Chat-v1.5~\cite{chen2024far} &\underline{57.1}	&46.8&	\underline{33.7}&	54.7&	 47.5&	80.6&	55.0 \\
\bf InternLM-XComposer2~\cite{internlmxcomposer2} &56.2	&41.4	&21.2&	\bf 59.5&	\bf 49.0&	\underline{81.2}&	\underline{57.5}\\
\bf LLaVA-NeXT (Yi-34B)~\cite{liu2024llavanext} &51.6&	\underline{48.8}&	26.7&	40.4&	29.8&	78.9&	51.6\\
\bf LLaVA-NeXT (Vicuna-13B)~\cite{liu2024llavanext} &40.4&	37.3&	16.5&	34.1&	19.6&	72.2&	36.2\\
\bf LLaVA-NeXT (Mistral-7B)~\cite{liu2024llavanext} &38.4&	37.0&	19.2&	34.6&	21.3&	69.0&	32.3\\
\bf LLaVA-NeXT (Vicuna-7B)~\cite{liu2024llavanext} &37.6&	37.6&	18.0&	31.5&	17.1&	67.0&	29.3 \\
\bf Yi-VL-34B~\cite{2023YiVL}	&40.5&	45.1&	21.2&	31.5&	12.0&	65.9&	26.4 \\
\bf Emu2-Chat~\cite{sun2023emu2chat}	&40.7&	35.0&	27.8&	30.7&	14.0	&49.7&	22.7 \\
\bf LLaVA-InternLM2-20B~\cite{2023xtuner}	&41.9&	39.4&	18.0&	25.3&	9.9&	65.4&	28.4\\
\bf DeepSeek-VL-7B~\cite{lu2024deepseek} & 40.5 & 38.3 & 19.6	& 36.9 & 20.7	& 65.3 & 36.9\\
\bf MiniCPM-V-2~\cite{minicpm2024}  &39.1&	38.2	&21.2&	39.8&	25.4&	62.9&	26.9 \\
\bf LLaVA-v1.5-13B~\cite{liu2023improved} &34.3	&37.0	&15.3&	27.7&	10.3&	61.1&	23.0 \\
\bf LLaVA-v1.5-7B~\cite{liu2023improved} &33.1&	35.7&	15.7&	25.6&	8.5&	55.5&	17.8\\
\bf mPLUG-Owl2~\cite{ye2023mplugowl2}	&34.8	&34.7	&19.6	&25.4	&8.5	&55.7	&20.5 \\
\rowcolor{t_green}\multicolumn{8}{c}{\emph{Open-Source VLMs (E2E Baselines)}}  \\
\bf LLaVA-InternLM2-1.8B~\cite{2023xtuner} &34.5	&30.2	&18.4	&\underline{26.3}&	\underline{9.1}&	43.6&	23.5 \\
\bf LLaVA-InternLM2-7B~\cite{2023xtuner} &\underline{38.3}	&\underline{40.1}&	\underline{21.9}&	26.0&	7.4&	\underline{63.6}&	\underline{26.6} \\
\rowcolor{orange!10}\multicolumn{8}{c}{\emph{Prism Models}}  \\
\bf ShareCaptioner-\texttt{ChatGPT}	&38.7	&45.2&	30.2&	30.2&	12.6&	59.6&	22.2\\
\bf PrismCaptioner-2B-\texttt{ChatGPT}	&43.3&	46.6&	32.0	&33.5&	18.2&	62.0	&27.4 \\
\bf PrismCaptioner-2B-\texttt{Llama3}	&42.0	&46.7	&32.4&	33.5&	19.8&	59.8&	30.3\\
\bf PrismCaptioner-7B-\texttt{ChatGPT}	&43.7&	47.3&	29.8&	35.1&	24.6&	65.4&	30.6\\
\bf PrismCaptioner-7B-\texttt{Llama3}	&\underline{45.9}&	\underline{53.3}&	\underline{35.6}&	\underline{39.0}&	\underline{27.3}&	\underline{68.1}&	\underline{37.2}\\
\bottomrule
\end{tabular}}
\caption{\textbf{Detail Results of Models under Prism Framework.}\protect\footnotemark (F) represents the sub-dataset filtered by our strategy in order to ensure vision indispensability and avoid data leakage. The suffix of Prism models (\texttt{ChatGPT} or \texttt{Llama3}) indicates the reasoning module adopted.}
% Models are sorted by the descending order of overall score with query-specific instruction. 
\label{tab:detail-performance-prism}
\end{table*}

\footnotetext{For comparison, we use single image as input and limit the maximum numbers of output tokens to 512. Better performance of \texttt{PrismCaptioners} is presented in \cref{sec:better-performance}}
\vskip -0.5em
\section{Discussion}
\vskip -0.5em
\textbf{Prism's Value as an Evaluation Framework. } 
Prism's value as an Evaluation Framework lies in its ability to 
disentangle and measure the perception and reasoning capabilities 
of VLMs across various data sources. 
There exist specialized multimodal benchmarks designed to assess VLMs' perception and reasoning capabilities, yet they often focus on specific domains.
For instance, 
RealWorldQA~\cite{RealWorldQA} evaluates real-world perception with high-resolution images, 
OCRVQA~\cite{mishra2019ocr} assesses text recognition in publications, 
and POPE~\cite{li2023evaluating} determines object existence in images.
However, 
many interested domains (geometry, medical images, GUIs, \textit{etc.}) are not covered by those perception benchmarks. 
Prism fills this gap by enabling the measurement and comparison of VLMs' perception capabilities on general VQA datasets in these domains. 
Additionally, existing `reasoning' benchmarks~\cite{lu2023mathvista,han2023infimm}
are compositional, 
requiring VLM to recognize key elements and before reasoning.
Comparing an end-to-end VLM with Prism equipped with the same VLM and an external LLM, like ChatGPT, 
provides insights into the VLM's intrinsic reasoning capabilities and potential performance constraints. 

\begin{wraptable}{r}{0.52\linewidth}
\vspace{-2mm}
\centering
\tablestyle{5pt}{1.3}
\resizebox{\linewidth}{!}{
\begin{tabular}{l|cccccc|c}
\toprule
% & MMStar & MMMU & \makecell{MMMU\\filter} & MathVista & \makecell{MathVista\\filter} & AI2D\_TEST & \makecell{AI2D\_TEST\\filter} \\
\bf VLM & \textbf {CP} & \textbf {FP} & \textbf {IR} & \textbf {LR} & \textbf {Math} & \textbf {ST} & \textbf {Overall} \\
\midrule
\bf GPT-4v &	67.2	&43.6	&56.8&	46.8&	42.0&	30.8&	47.9  \\
\bf GeminiPro-V &61.2&	35.2&	53.2&	47.6&	48.4&	19.2&	44.1 \\
\midrule
\bf Ensemble & 66.4	& \bf 47.2$\uparrow$ &	\bf 58.8$\uparrow$ &  \bf 50.8$\uparrow$ &	46.4&	\bf 32.0$\uparrow$&	\bf 50.3$\uparrow$ \\
\bottomrule
\end{tabular}
}
\caption{\textbf{Performance of Perception Module with Multiple VLMs on MMStar.}}
\vskip -1.0em
\label{tab:wrap_merged_desc}
\end{wraptable} 

\textbf{Prism's Value as a Vision-Language Solver. } 
By integrating small-scale VLMs with readily available external LLMs within Prism, we achieve superior performance compared to the standard end-to-end capabilities of the standalone VLM. This approach also renders it practical to address vision-language tasks using a 2B parameter VLM,
as reasoning is effectively outsourced to external LLMs. 
When implemented with an LLM API, 
Prism's inference process (without quantization) only consumes several gigabytes of GPU memory. 
Furthermore, Prism allows for the flexible incorporation of multiple VLMs to enhance perception.
For instance, the straightforward concatenation of outputs from GPT-4v and GeminiPro-V has demonstrated substantial improvements across the majority of metrics on the MMStar benchmark, 
as substantiated by the data presented in Table \ref{tab:wrap_merged_desc}.

% \input{tables/merged_description_performance}

% \textbf{Conclusion. } We introduce a novel framework Prism for explicitly disentangling the perception and reasoning processes. It can serve as both an evaluation framework and a 
% vision-language solver. Utilizing Prism, we conduct a comprehensive analysis of perception and reasoning capabilities of VLMs and bring up some intriguing findings. Moreover, Prism provides a prospective way of using external LLMs to substitute for VLMs when reasoning. we explore the requisite for a powerful visual captioner, and verify the potential of Prism by combining a 2B-parameter vision captioner and ChatGPT-3.5, which achieves fabulous performance on MMStar and other domain-specific benchmarks. We hope this work could contribute to the development of the VLM community.

% \hdc{If there is space left, we can also move Limitations and Broader Impacts Here. }
\section{Limitations and Broader Impacts}
\label{sec:impacts}

\textbf{Limitations. }
In this study, we introduce the Prism framework and 
showcase its effectiveness as both an analytical tool 
and a versatile vision-language task solver. 
Given budget constraints, 
our evaluation focuses on a select group of representative open-source and proprietary VLMs, 
which may not encompass all the most recent high-performing models. 
Our experiments with training visual captioners aim to illustrate Prism's capability to deliver strong performance on vision-language tasks while minimizing costs. 
To fully realize the potential of Prism,
additional experiments are recommended, 
particularly on domain-specific visual instruction tuning data, 
such as tables and diagrams, screens, and graphical user interfaces (GUIs), 
medical images, etc., 
to thoroughly assess Prism's efficacy in these specialized contexts.

\textbf{Broader Impacts. }
As an analytical framework, 
Prism offers detailed insights into the perception and reasoning capabilities of vision-language models (VLMs), 
providing valuable guidance for future model optimization. 
Training large-scale VLMs necessitates extensive multi-modal data and computational resources.
Prism mitigates this challenge by training small-scale VLMs that specialize in visual captioning tasks and perform reasoning with large language models (LLMs), which are now cost-effective.\footnote{Thanks to the development in deploying technologies, the inference of LLM is now cheap, provided by various corporations at a price as low as millions of tokens per US dollar. }
When employed as a vision-language task solver, 
Prism can be trained and deployed at a significantly reduced cost, 
making it a promising approach for highly customized applications or tasks with limited training data.
However, Prism also carries potential societal impacts,
as it could lower the barrier to building multimodal applications, some of which may be harmful. 
There is a risk that Prism could be used to develop harmful multimodal AI systems. 
Additionally, data-driven methods often inherit biases, which can persist in downstream tasks. 
We urge users to thoughtfully consider the implications of these biases when implementing our model.

\appendix
\section{Related Work}

% 随着大模型的不断发展，人们开始探索将多模态模块和大语言模型进行结合，从而让语言模型能通过视觉感知来增强与现实世界的结合。[kosmos-2, minigpt-4, blip2]。blip系列和Visual ChatGPT分别利用了q-former和foundation model来证明了不同模态间可进行特征对齐以完成Zero-shot Visual question and answering等任务。llava和minigpt-4的出现，使得使用简单的 MLP-based 衔接层以弥合模态差异成为了可能。在这一框架中，视觉模型和衔接层的选用，成为了各个多模态模型的性能提升焦点。大部分MLLM采用CLIP模型（如llava, qwen-vl等）因其在预训练过程中已将视觉与文本对齐，部分则采用经过优化的CLIP变种模型以期望提升性能及训练效率，如 MiniGPT-4采用EVA-CLIP。而随着高分辨率图片的处理要求不断出现，部分模型采用hybrid vision encoder（如deepseek-vl) 从而对细节及关键视觉特征进行更好的捕获。在衔接层的选择上，较为轻量化且广泛采用的方案为使用 MLP-based 衔接层（如llava,llava-next），而qwen-vl，cogvlm等方案通过额外的注意力层以实现文本特征与视觉特征之间的深度交互和融合，并通过可学习的提示来使得语言模型能够聚焦于更核心的视觉特征。

\subsection{Large Vision-Language Models (LVLMs)}

The landscape of Large Language Models (LLMs) is continually evolving, 
with an expanding body of research focused on integrating multimodal capabilities to enhance their perceptual abilities in real-world contexts \cite{peng2023kosmos,zhu2023minigpt,li2023blip}.
Early efforts in this direction, 
such as Flamingo~\cite{alayrac2022flamingo},
introduced gated dense blocks of cross-attention within pre-trained language encoder layers to fuse visual features.
Subsequent models like BLIP2~\cite{li2023blip} and InstructBLIP~\cite{dai2023instructblip} 
utilized a Q-former to align features across different modalities, 
enabling tasks such as zero-shot visual question answering.
More recent models, including LLaVA~\cite{liu2023visual} and MiniGPT-4~\cite{zhu2023minigpt}, 
have simplified modality bridging through the use of MLP-based projection layers, d
offering a more straightforward approach compared to the Q-former.
The architecture of LLaVA has been widely adopted in subsequent works~\cite{liu2024llavanext,2023YiVL,chen2024far,lu2024deepseek}. 
The choice of vision encoders and the language model is considered critical for the overall performance of VLMS. 
Most VLMs 
\cite{liu2023visual,internlmxcomposer2,liu2024llavanext,laurencon2023obelics,zhu2023minigpt} 
employ CLIP-based Vision Transformer~\cite{radford2021learning,zhai2023sigmoid,sun2023eva} as the vision encoder, 
owing to its good pre-training alignment of visual and textual modalities. 
There is a prevalent belief~\cite{liu2023mmbench,liu2024llavanext} that the scale of the language model significantly impacts the performance of VLMs, though detailed analyses are lacking.
In addition to the open-source VLMs developed by the academic community, 
numerous proprietary VLMs~\cite{OpenAI2023GPT4TR,team2023gemini,bai2023qwen,ormazabal2024reka} demonstrate robust performance across various multimodal benchmarks. 
This paper presents a breakdown capability analysis of both open-source and proprietary VLMs using the Prism framework.

% With the growing need to process high-resolution images, some models have adopted hybrid vision encoders (such as DeepSeek-VL\cite{lu2024deepseek}) to capture detailed and critical features better. 
% In terms of adapter layer selection, a widely adopted and lightweight approach is using MLP-based adapter layers (such as LLaVA\cite{liu2023visual}, 
% LLaVA-NeXT\cite{liu2024llavanext}). 
% Meanwhile, models like Qwen-VL\cite{bai2023qwen} and CogVLM\cite{Wang2023CogVLMVE} incorporate additional attention layers to deeply integrate and interact with text and visual features, 
% using learnable prompts to focus the language model on more central visual elements.

\subsection{LVLMs Capability Evaluation}

Large-scale VLMs have demonstrated promising outcomes across a diverse range of multimodal tasks, 
as evidenced by extensive qualitative and quantitative evaluations. 
Early assessments of LVLMs often involved open-ended Visual Question Answering (VQA)
\cite{nocaps,hudson2019gqa,ok-vqa,vqav2} 
and human-based subjective evaluations~\cite{ye2023mplug,Xu2023LVLMeHubAC}.
However, these methods face limitations in accurately reflecting the true performance of VLMs. 
Open-ended VQA tasks typically demand an exact match between the model's prediction and the ground truth, 
which can lead to a significant number of false positives. 
Conversely, 
subjective evaluations introduce biases and make the results challenging to reproduce. 
Subsequent research has shifted towards structuring visual questions in closed-ended formats, 
such as multiple-choice or Yes-or-No questions.
Pioneering works like MMBench~\cite{liu2023mmbench}, MME~\cite{Fu2023MMEAC}, or SEEDBench~\cite{li2023seed} have presented comprehensive evaluations of VLMs using closed-ended VQA, covering various perception and reasoning capabilities.
Additionally, specialized multimodal benchmarks have emerged to assess VLMs from specific angles.
For instance, 
MMMU~\cite{yue2023mmmu} evaluates VLMs' ability to handle multimodal examination questions, 
while POPE~\cite{li2023evaluating} and HallusionBench~\cite{liu2023hallusionbench} 
scrutinize hallucination and illusion phenomena in VLMs.
RealWorldQA~\cite{RealWorldQA} focuses on real-world perception with high-resolution images.
Recently, MMStar~\cite{chen2024we} has addressed the issues of vision dispensability and data contamination in existing benchmarks 
by compiling high-quality, vision-indispensable questions from multiple sources, 
ensuring minimal data leakage and covering six core capabilities. 
In this study, Prism primarily utilizes MMStar for capability evaluation.

% As a comprehensive disciplinary LVLM evaluation benchmark, MMMU\cite{yue2023mmmu} contains approaches to assessing visual perception and discipline-specific reasoning abilities. 
% Models as powerful as GPT4V\cite{OpenAI2023GPT4TR} still commit basic perceptual errors and domain-specific perceptual errors in up to 35$\%$ of assessments. 
% Similarly, in MMBench\cite{liu2023mmbench}, Perception and Reasoning were adopted as VLM's level-1 abilities and subdivided into multiple fine-grained capabilities. 
% Both proprietary and open-source VLMs suffer from various capability deficiencies in MMBench\cite{liu2023mmbench}, 
% such as coarse-grained perception, which affects the overall performance of the models. 
% MMStar\cite{chen2024we} is a highly visual content-dependent and low-data-leakage multimodal evaluation benchmark that divides the LVLM into six core capabilities that enable a perfect assessment of the overall real-world performance of multimodal models. 
% Under this benchmark, LVLMs struggle to pass in core capabilities such as fine-grained perception (FP) and logical reasoning (LR). 
% Based on MMStar\cite{chen2024we}'s well-established ability categorization, combined with \textbf{Prism}, we can well evaluate the real visual perception ability of VLM.

\subsection{LVLMs Capability Breakdown Investigation}

To offer insightful and detailed feedback for the future optimization of VLMs,
some researchers have initiated efforts to dissect the abilities of VLMs, 
aiming to uncover strategies for enhancement.
To more effectively explore the disparity between VLMs and human cognition, 
Zhang et al.~\cite{zhang2024far} employ Raven's Progressive Matrices (RPM) to examine the model's deductive reasoning skills grounded in visual perception. 
Through error case analysis, 
the authors observe that VLMs often make compounding and confounding errors when articulating individual elements within RPM, 
which subsequently results in erroneous reasoning. 
Although this study provides qualitative insights, 
it does not encompass a systematic evaluation of perception and reasoning capabilities. 
Meanwhile, researchers have developed specialized benchmarks to assess specific capabilities. 
For example, InfiMM-Eval~\cite{han2023infimm} and MathVista~\cite{lu2023mathvista} have implemented rigorous, step-by-step evaluations of the model's complex reasoning abilities on natural images and mathematical VQA problems, respectively. 
However, these reasoning benchmarks require a foundational perception capability to accurately identify key elements. 
Concurrently, there are perception benchmarks \cite{tong2024eyes,RealWorldQA,li2023evaluating,fu2024blink}
that exclusively focus on evaluating the perception skills of VLMs across various scenarios. 
In this study, 
Prism introduces a general decoupling framework that facilitates a detailed analysis of perception and reasoning capabilities, applicable to any multimodal benchmark.

% while HallusionBench~\cite{li2023evaluating} , wang2023makes} investigated the limitations of the visual module on the model in terms of perceptual capabilities.
% the examination of fine-grained perception capabilities even involves multiple instances, and many LVLMs are also suffering from defective capacity and hallucination.
%the key challenges posed by MMLU encompass the requirement for expert visual perceptual skills and strong reasoning with discipline-specific knowledge.

%为了更好的将视觉模块和大语言模型进行对齐而不损失过多的视觉特征，大部分multimodal large language models（即MLLM）采用的均为在视觉模块和自回归模型间加入衔接层并加以多阶段训练，使得语言模型能与感知到的视觉模块进行语义对齐，从而作出精准的回答。由于对视觉特征的复杂性的看法存在一定差异，在衔接层的选择上，目前可大致分为最简单的linear MLP[llava相关论文, ]，改良后的transformer模块（如Q-former[blip-2], Perceiver Resampler[Flamingo]) 和 专家模型（如captioner[“Caption anything: Interactive image description with diverse multimodal control])。

% 为了增强
\section{Supplementary Details of Prism}

\subsection{Details of Prism Evaluation Framework}
\label{sec:evalution-details}

\subsubsection{Query-Specific Instruction Details}
% To make the query-specific part of the \textit{\colorbox{lightblue}{Query-Specific Instruction}} closely related to the question and options, we adopt the method of few-shot learning. For each of the questions in MMStar, we gave demanding instructions and used several different examples to enable the reasoning module to learn the meaning of "contents to observe" in these contexts and to make inferences in response to the specific questions, as shown in Figure 10 for the prompts used.

\textit{\colorbox{lightblue}{Query-Specific Instruction}} is a combination of generic instruction and query-specific part, 
as depicted in \cref{fig: instructions}. 
To ensure that the query-specific part generated by the reasoning module is closely related to the questions and options, 
we adopt few-shot learning to guide the LLM. 
For each visual question, 
we feed the request with multiple examples. 
This approach helps the reasoning module understand what ``contents to observe" means in different contexts and allows it to make accurate inferences in response to specific questions, as illustrated by the prompts in \cref{fig: few-shot-prompt}.

\begin{figure*}[!ht] 
% \vspace{-5mm}
\begin{AIbox}{Instructions}
{
    \textbf{\textcolor{blue}{Generic Instruction: }}Describe the fine-grained content of the image, including scenes, objects, relationships, instance location, and any text present.
    
    \textbf{\textcolor{blue}{Query-Specific Instruction: }}Describe the fine-grained content of the image, including scenes, objects, relationships, instance location, and any text present. Especially, pay attention to \textcolor{brown}{<query-specific part>}.
    
}
\end{AIbox} 
\caption{\textbf{Generic Instruction \textit{vs.} Query-Specific Instruction. }}
\label{fig: instructions}
% \vspace{-5mm}
\end{figure*}

\begin{figure*}[!ht] 
% \vspace{-5mm}
\begin{AIbox}{The Few-shot Prompt Template}
{
    Your task is to give a concise instruction about what basic elements are needed to be described based on the given question. Ensure that your instructions do not cover the raw question, options, or thought process of answering the question. \\
    \\
    \textbf{\textcolor{blue}{Question}:} In which period the number of full-time employees is the maximum? \\
    \textbf{\textcolor{blue}{Contents to observe:}} the number of full-time employees\\
    \textbf{\textcolor{blue}{Question}:} What is the value of the smallest bar?\\
    \textbf{\textcolor{blue}{Contents to observe:}} the heights of all bars and their values\\
    \textbf{\textcolor{blue}{Question}:} What is the main subject of the image?\\
    \textbf{\textcolor{blue}{Contents to observe:}} the central theme or object\\
    \textbf{\textcolor{blue}{Question}}: What is the position of the catcher relative to the home plate?\\
    \textbf{\textcolor{blue}{Contents to observe:}} the spatial arrangement of the objects \\
    \textbf{\textcolor{blue}{Question}:} What is the expected ratio of offspring with white spots to offspring with solid coloring? Choose the most likely ratio.\\
    \textbf{\textcolor{blue}{Contents to observe:}} the genetic information\\
    \textbf{\textcolor{blue}{Question}:} \textcolor{brown}{<question>}\\
    \textbf{\textcolor{blue}{Contents to observe:}}\\
}
\end{AIbox} 
\caption{\textbf{The Prompt Template for the Reasoning Module to Generate the "Contents to Observe" Part.}}
\label{fig: few-shot-prompt}
% \vspace{-5mm}
\end{figure*}

\subsubsection{Inference Prompt Template of the Reasoning Module}
%在感知模块产生了对画面具体描述后，需要对query进行改写以让推理模块能根据描述并结合问题进行更加准确的作答。改写后的query的模板如图12所示，此处的改写仅仅采用了简单拼接的方式，直观的让推理模块进行直接作答。示例可见图13。
% After the perception module produces a specific description of the image, the query needs to be rewritten to allow the reasoning module to answer more accurately based on the description and the question. The template of the rewritten query is shown in Fig. 12, where the rewriting is done in a simple splicing manner, which is intuitive for the reasoning module to answer directly. An example can be seen in Figure 13.
After the perception module generates detailed visual information about the image, 
the query needs to be reformatted to enable the reasoning module to answer more accurately based on the information and the question. 
The template for reformatting is shown in \cref{fig: rewritten-query-template}. The reformatting process involves simple splicing, making it intuitive for the reasoning module to respond directly. An example of this can be seen in \cref{fig: rewritten-query-example}.

\begin{figure*}[!ht] 
% \vspace{-5mm}
\begin{AIbox}{Reformatting Template}
{
    You are an excellent text-based reasoning expert. You are required to answer the question based on the detailed description of the image.\\
    \textbf{\textcolor{blue}{Description:}} \textcolor{brown}{<description>} \\
    \textbf{\textcolor{blue}{Question:}} \textcolor{brown}{<question>}
}
\end{AIbox} 
\caption{\textbf{The Template for Reformatting Query.}}
\label{fig: rewritten-query-template}
% \vspace{-5mm}
\end{figure*}

\begin{figure*}[!ht] 
% \vspace{-5mm}
\begin{AIbox}{Reformatted Query Example}
{
    You are an excellent text-based reasoning expert. You are required to answer the question based on the detailed description of the image.\\
    \textbf{\textcolor{blue}{Description:}} The image presents a delightful celebration of Father's Day. Dominating the center of the image is a blue tie, adorned with white stripes, symbolizing the essence of fatherhood. The tie is slightly tilted to the right, adding a touch of dynamism to the composition. 
    On the left side of the tie, the phrase "Happy Father's Day" is elegantly inscribed in a white cursive font, extending warm wishes to all dads. The text and the tie are set against a dark blue background, creating a striking 
    contrast that draws attention to the main elements of the image. 
    Adding a final touch of sophistication, a thin white border frames the entire image, encapsulating the joyous message of Father's Day. The image, in its entirety, serves as a heartfelt tribute to all the wonderful fathers out 
    there.
    
    \textbf{\textcolor{blue}{Question:}} Which special day is associated with this poster?
    
    Options:
    A. Earth Day.
    B. National Reading Day.
    C. Father's Day.
    D. Mother's Day
    
    Please select the correct answer from the options above. \\
    
}
\end{AIbox} 
\caption{\textbf{An Example of the Reformatted Query. }}
\label{fig: rewritten-query-example}
% \vspace{-5mm}
\end{figure*}

\subsection{More Details of PrismCaptioner Training}
\label{sec:lora-details}

The details about QLoRA for training PrismCaptioner is demonstrated in \cref{tab:PrismCaptioner-train-detail}

\begin{table*}[t]
\vspace{-5mm}
\centering
% \vskip -1.1em
% Result of Instruction\emph{Query-Specific Instruction} and \emph{Generic Instruction}
\resizebox{0.7\linewidth}{!}{
\tablestyle{36pt}{1.4}
\begin{tabular}{cc}
\toprule
\bf Hyperparameter & \bf Assignment  \\
\midrule
\bf Image Resolution   & 384 * 384 \\
\bf Patch Size & 14  \\
\bf Max Length & 1296 \\
% \bf evaluation frequency & 500 iterations \\
\midrule
\bf QLoRA Quantization & 4bit \\
\bf LLM int8 Threshold & 6.0 \\
\bf BnB 4bit Compute Dtype & torch.float16 \\
\bf BnB 4bit Double Quant & True \\
\bf BnB 4bit Quant Dtype & nf4 \\
\midrule
\bf LoRA r & 512 \\
\bf LoRA $\alpha$ & 256 \\
\bf LoRA Dropout & 0.05 \\
\bottomrule
\end{tabular}
}
\vspace{2mm}
\caption{\textbf{PrismCaptioner Training Details.}}
\label{tab:PrismCaptioner-train-detail}
\end{table*}

\subsection{More Performance Details of PrismCaptioner}
\label{sec:better-performance}

In addition to the results of \cref{tab:detail-performance-prism}, we use multiple images as iputs if there are and set maximum output length to 2048. The performace is presented in \cref{tab:better-performance-prism}
\begin{table*}[t]
\centering
% \vskip -1.1em
% Result of Instruction\emph{Query-Specific Instruction} and \emph{Generic Instruction}
\resizebox{1.0\linewidth}{!}{
\tablestyle{7pt}{1.4}
\begin{tabular}{l|c|cc|cc|cc}
\toprule
\bf Model & \textbf{MMStar} & \textbf{MMMU} & \makecell{\textbf{MMMU (F)}} & \textbf{MathVista} & \makecell{\textbf{MathVista (F)}} & \textbf{AI2D} & \makecell{\textbf{AI2D (F)}} \\
\midrule
\rowcolor{orange!10}\multicolumn{8}{c}{\emph{Prism Models (Multiple Image Inputs \& Max Output Length 2048)}}  \\
\bf PrismCaptioner-2B-\texttt{ChatGPT}	&43.0	& 46.1	& 31.4	& 34.8	& 16.7	& 62.1	& 27.9 \\
\bf PrismCaptioner-2B-\texttt{Llama3}	&42.6	& 49.0 & 34.1	& 35.1	& 19.8	& 59.8	& 30.3\\
\bf PrismCaptioner-7B-\texttt{ChatGPT}	& 43.4	& 47.9	& 33.7	& 36.5	& 25.0 & 65.4	& 30.8\\
\bf PrismCaptioner-7B-\texttt{Llama3}	&\underline{46.2}&	\underline{56.5}&	\underline{42.0}&	\underline{39.8}&	\underline{26.2 }&	\underline{67.9}&	\underline{37.7}\\
\bottomrule
\end{tabular}}
\caption{\textbf{More Detailed Performance Results of PrismCaptioners}}
% Models are sorted by the descending order of overall score with query-specific instruction. 
\label{tab:better-performance-prism}
\end{table*}

% \section{Details for Prism Captioner Training}

\section{Detailed Examples of Prism}
\label{sec:examples}

\subsection{End-to-end \textit{v.s.} Prism Predictions}
\vspace{-1mm}
Some end-to-end models with small language models, often predict incorrect answers due to their limited reasoning capabilities. Utilizing an external LLM, Prism endows VLMs with the reasoning ability to solve the vision-language tasks, as presented in \cref{tab:case7} and \cref{tab:case8}.

\subsection{PrismCaptioner Performances}
\vspace{-1mm}
PrismCaptioner can professionally extract and express detailed visual information to solve coarse perception, fine-grained perception, instance reasoning, logical reasoning, science \& technology, and math tasks, as presented in \cref{tab:case1,tab:case2,tab:case3,tab:case4,tab:case5,tab:case6}.

\subsection{Performances Comparison between PrismCaptioner and GPT-4o}
Although PrismCaptioner can generate detailed descriptions of images, there is still room for improvement compared to GPT-4o.

(1) GPT-4o generates descriptions that are more relevant to the questions. Due to limited training data, PrismCaptioner's captions are less adaptive to query-specific parts compared to those of GPT-4o, as shown in \cref{tab:case11}.

(2) GPT-4o's expression is more detailed and specific, and the responses about spatiotemporal information are also more accurate, as shown in \cref{tab:case12}.

\begin{table}[t]
\begin{tcolorbox}[colback=gray!10, colframe=black, arc=5mm, boxrule=0.5mm, width=\textwidth]
\begin{tabularx}{\textwidth}{X}
\rowcolor{gray!10} 
\textbf{Question:} \textbf{What is the primary scene depicted in the image?} \\
\rowcolor{gray!10}
\textbf{Image:}\\
\rowcolor{gray!10}
\begin{center}
\includegraphics[width=0.55\textwidth]{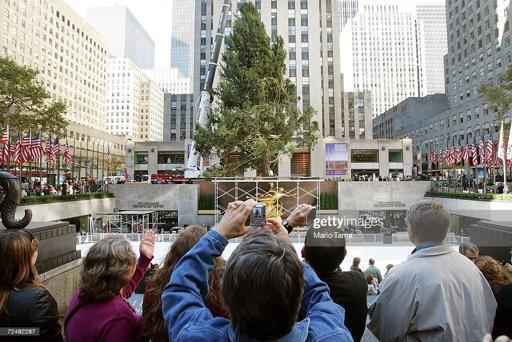}
\end{center} \\
\rowcolor{gray!10}
\textbf{Options:}
\newline A. A group of people taking pictures of a statue in a plaza
\newline B. A large crowd taking pictures of each other in an outdoor event
\newline C. A group of people taking pictures of a Christmas tree in a plaza
\newline D. A group of people taking pictures of a famous building in New York City\\
\rowcolor{gray!10}
\textbf{Category:} Coarse Perception \\
\rowcolor{gray!10}
\textbf{Model:} \texttt{LLaVA-NeXT (Vicuna-7B)} \\
\rowcolor{gray!10}
\textbf{End-to-end Prediction:}  D. A group of people taking pictures of a famous building in New York City ( \textcolor{red}{\faTimes} )\\
\rowcolor{gray!10}
\textbf{Perception Description:} \textcolor{blue}{The image captures a bustling scene in \textbf{a city square}, where \textbf{a large crowd of people} has gathered.} The crowd is diverse, with individuals of various ages and attire, all seemingly engaged in the event taking place. \textcolor{blue}{The central focus of the image is a large, \textbf{illuminated Christmas tree}, which stands prominently in the square.} The tree is adorned with a multitude of colorful lights and ornaments, creating a festive atmosphere.
\newline In the background, the cityscape is visible with tall buildings, some of which are adorned with American flags, suggesting a location in the United States. The flags are flying at half-mast, adding a solemn tone to the scene.
\newline The sky above is overcast, casting a soft light over the scene. The overall mood of the image is one of celebration and unity, as the crowd seems to be enjoying the holiday festivities. The image does not contain any text. The relative positions of the objects suggest a well-organized event, with the Christmas tree as the focal point and the crowd surrounding it. The flags in the background provide a sense of location and national pride. The overcast sky adds a touch of solemnity to the otherwise joyful scene.\\
\rowcolor{gray!10}
\textbf{Perception Prediction:}  C. ( \textcolor{darkgreen}{\faCheck} )\\
\end{tabularx}
\end{tcolorbox}
\caption{\textbf{An Example of End-to-End \textit{v.s.} Prism \texttt{LLaVA-NeXT (Vicuna-7B)} Prediction.}}
\label{tab:case7}
\end{table}

\begin{table}[t]
\begin{tcolorbox}[colback=gray!10, colframe=black, arc=5mm, boxrule=0.5mm, width=\textwidth]
\begin{tabularx}{\textwidth}{X}
\rowcolor{gray!10} 
\textbf{Question:} \textbf{Which number is missing?} \\
\rowcolor{gray!10}
\textbf{Image:}\\
\rowcolor{gray!10}
\begin{center}
\includegraphics[width=0.55\textwidth]{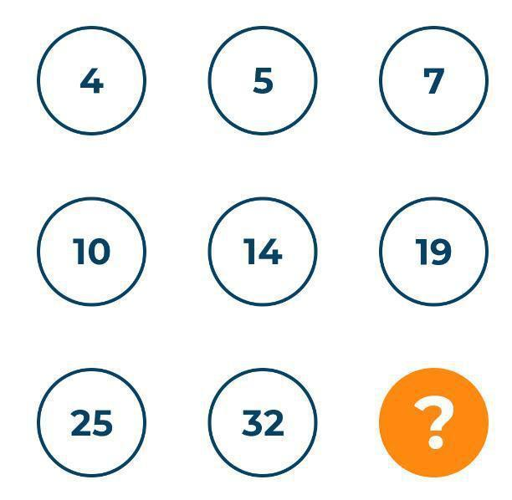}
\end{center} \\
\rowcolor{gray!10}
\textbf{Options:}
A. 40 \quad B. 10 \quad C. 22 \quad D. 34\\
\rowcolor{gray!10}
\textbf{Category:} Logical Reasoning \\
\rowcolor{gray!10}
\textbf{Model:} \texttt{MiniCPM-V-2} \\
\rowcolor{gray!10}
\textbf{End-to-end Prediction:}  B. 10 ( \textcolor{red}{\faTimes} )\\
\rowcolor{gray!10}
\textbf{Perception Description:} The image presents a grid of nine circles, each containing a number from 4 to 25. The numbers are arranged in a 3x3 grid, with the numbers 4, 5, 7, 10, 14, 19, 25, 32, and a question mark (?) at the bottom right corner. \textcolor{blue}{The sequence of numbers in the grid is 4, 5, 7, 10, 14, 19, 25, 32, with the question mark indicating the missing number.}\\
\rowcolor{gray!10}
\textbf{Perception Prediction:} The pattern in the sequence is as follows:
5 - 4 = 1, 7 - 5 = 2, 10 - 7 = 3, 14 - 10 = 4, 19 - 14 = 5, 25 - 19 = 6, 32 - 25 = 7. Therefore, the missing number should be 32 + 8 = 40.
\newline So, the correct answer is: A. 40 ( \textcolor{darkgreen}{\faCheck} )\\
\end{tabularx}
\end{tcolorbox}
\caption{\textbf{An Example of End-to-End \textit{v.s.} Prism \texttt{MiniCPM-V-2} Prediction.}}
\label{tab:case8}
\end{table}

\begin{table}[t]
\begin{tcolorbox}[colback=gray!10, colframe=black, arc=5mm, boxrule=0.5mm, width=\textwidth]
\begin{tabularx}{\textwidth}{X}
\rowcolor{gray!10} 
\textbf{Question:} \textbf{Which image is more brightful?} \\
\rowcolor{gray!10}
\textbf{Image:}\\
\rowcolor{gray!10}
\begin{center}
\includegraphics[width=0.65\textwidth]{figs/case3.png}
\end{center} \\
\rowcolor{gray!10}
\textbf{Options:} A. The first image \quad B. The second image \\
\rowcolor{gray!10}
\textbf{Category:} Coarse Perception\\
\rowcolor{gray!10}
\textbf{Description:} The image is a split composition with two distinct halves.
\newline \textcolor{blue}{On the left side, there is a photograph of a \textbf{bright} red sports car parked on the side of a road.} The car appears to be a modern coupe with a sleek design, featuring a long hood and a short rear deck, which are characteristic of high-performance vehicles. It has five-spoke alloy wheels with a silver finish and low-profile tires. The car's windows are tinted, and it has a visible antenna on the rear part of the roof. The background shows a blurred natural landscape with greenery, suggesting a rural or semi-rural setting. 
\newline \textcolor{blue}{On the right side of the image, there is a digital artwork depicting a swirling, fractal-like pattern in shades of green and purple, set against a \textbf{dark} background.} The pattern has a hypnotic, almost hypnotic quality, with a sense of motion and energy emanating from it. The artwork has a hypnotic, almost hypnotizing effect, drawing the viewer's attention to the center of the swirl where the colors seem to converge.\\
\rowcolor{gray!10}
\textbf{Prediction:}  A. The first image ( \textcolor{darkgreen}{\faCheck} )\\
\end{tabularx}
\end{tcolorbox}
\caption{\textbf{PrismCaptioner Generates Description to Solve Coarse Perception Task.}}
\label{tab:case1}
\end{table}

\begin{table}[t]
\begin{tcolorbox}[colback=gray!10, colframe=black, arc=5mm, boxrule=0.5mm, width=\textwidth]
\begin{tabularx}{\textwidth}{X}
\rowcolor{gray!10} 
\textbf{Question:} \textbf{What color is the shirt of the man playing the guitar?} \\
\rowcolor{gray!10}
\textbf{Image:}\\
\rowcolor{gray!10}
\begin{center}
\includegraphics[width=0.65\textwidth]{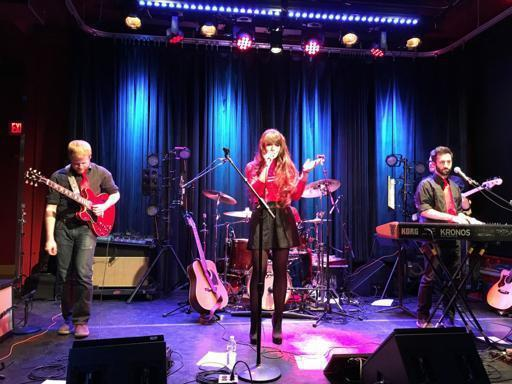}
\end{center} \\
\rowcolor{gray!10}
\textbf{Options:} A. Blue \quad B. Brown \quad C. Red \quad D. Black \\
\rowcolor{gray!10}
\textbf{Category:} Fine-grained Perception \\
\rowcolor{gray!10}
\textbf{Description:} The image depicts a live music performance by a band on stage. There are three members visible: a vocalist, a guitarist, and a keyboard player. The vocalist, positioned in the center, is a woman with long hair, wearing a red top and a black skirt. She is holding a microphone stand with her right hand and appears to be singing into it. \textcolor{blue}{To her left, a man is playing an electric guitar, dressed in a \textbf{black shirt} and jeans, focused on his instrument}. To the right of the vocalist, another man is playing a keyboard, wearing a black shirt and dark pants. The stage is equipped with various musical instruments and equipment, including a drum set behind the vocalist, a keyboard in front of the keyboard player, and a microphone stand in front of the vocalist. The lighting is dim, with spotlights highlighting the performers, and the background is a draped curtain in shades of blue and red. The atmosphere suggests an intimate concert setting.\\
\rowcolor{gray!10}
\textbf{Prediction:}  D. Black ( \textcolor{darkgreen}{\faCheck} )\\
\end{tabularx}
\end{tcolorbox}
\caption{\textbf{PrismCaptioner Generates Description to Solve Fine-Grained Perception Task.}}
\label{tab:case2}
\end{table}

\begin{table}[t]
\begin{tcolorbox}[colback=gray!10, colframe=black, arc=5mm, boxrule=0.5mm, width=\textwidth]
\begin{tabularx}{\textwidth}{X}
\rowcolor{gray!10} 
\textbf{Question:} \textbf{Which solution has a higher concentration of pink particles?} \\
\rowcolor{gray!10}
\textbf{Image:}\\
\rowcolor{gray!10}
\begin{center}
\includegraphics[width=0.5\textwidth]{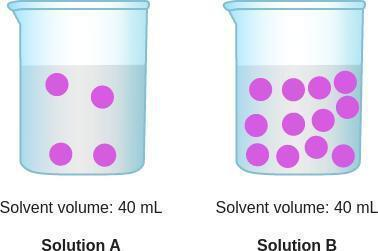}
\end{center} \\
\rowcolor{gray!10}
\textbf{Options:} A. neither; their concentrations are the same \quad B. Solution A \quad C. Solution B \\
\rowcolor{gray!10}
\textbf{Category:} Instance Reasoning\\
\rowcolor{gray!10}
\textbf{Description:} The image displays two beaker-like containers, labeled Solution A and Solution B, each containing a clear liquid. Solution A has a volume of 40 mL, while Solution B has a volume of 40 mL. Inside Solution A, there are multiple small, round, purple particles dispersed uniformly throughout the liquid. In contrast, \textcolor{blue}{Solution B contains a similar number of purple particles, but they are clustered \textbf{more densely}, suggesting a higher concentration}. The background is plain white, focusing attention on the containers and their contents. There are no additional background elements or context provided.\\
\rowcolor{gray!10}
\textbf{Prediction:}  C. Solution B ( \textcolor{darkgreen}{\faCheck} )\\
\end{tabularx}
\end{tcolorbox}
\caption{\textbf{PrismCaptioner Generates Description to Solve Instance Reasoning Task.}}
\label{tab:case3}
\end{table}

\begin{table}[t]
\begin{tcolorbox}[colback=gray!10, colframe=black, arc=5mm, boxrule=0.5mm, width=\textwidth]
\begin{tabularx}{\textwidth}{X}
\rowcolor{gray!10} 
\textbf{Question:} \textbf{In nature, what's the relationship between these two creatures?} \\
\rowcolor{gray!10}
\textbf{Image:}\\
\rowcolor{gray!10}
\begin{center}
\includegraphics[width=0.65\textwidth]{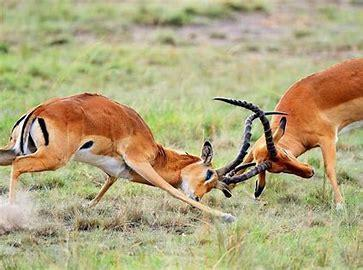}
\end{center} \\
\rowcolor{gray!10}
\textbf{Options:} 
\newline A. Predatory relationships \quad B. Competitive relationships
\newline C. Parasitic relationships \quad D. Symbiotic relationship \\
\rowcolor{gray!10}
\textbf{Category:} Logical Reasoning\\
\rowcolor{gray!10}
\textbf{Description:} \textcolor{blue}{The image depicts two antelopes engaged in what appears to be \textbf{a physical confrontation} on a grassy savanna.} The antelope on the left is rearing up on its hind legs, with its front legs extended towards the other antelope, which is also on its hind legs, facing the first antelope. Both animals have prominent, curved horns, and their fur is a rich, reddish-brown color. The background is a blurred natural landscape, suggesting a wild, open environment typical of a savanna ecosystem. There are no human elements or artificial structures visible, reinforcing the natural setting. The colors are vibrant, with the green of the grass contrasting with the reddish-brown of the antelopes.\\
\rowcolor{gray!10}
\textbf{Prediction:}  B. Competitive relationships ( \textcolor{darkgreen}{\faCheck} )\\
\end{tabularx}
\end{tcolorbox}
\caption{\textbf{PrismCaptioner Generates Description to Solve Logical Reasoning Task.}}
\label{tab:case4}
\end{table}

\begin{table}[t]
\begin{tcolorbox}[colback=gray!10, colframe=black, arc=5mm, boxrule=0.5mm, width=\textwidth]
\begin{tabularx}{\textwidth}{X}
\rowcolor{gray!10} 
\textbf{Question:} \textbf{What comes after the full moon?} \\
\rowcolor{gray!10}
\textbf{Image:}\\
\rowcolor{gray!10}
\begin{center}
\includegraphics[width=0.46\textwidth]{figs/case7.png}
\end{center} \\
\rowcolor{gray!10}
\textbf{Options:} A. waning cresent \quad B. 3rd quarter \quad C. waning Gibbous \quad D. new moon \\
\rowcolor{gray!10}
\textbf{Category:} Science \& Technology \\
\rowcolor{gray!10}
\textbf{Description:} The image depicts a simplified diagram illustrating the phases of the Moon in relation to the Earth and the Sun. It is a circular diagram with the Sun at the top right and the Earth at the bottom left. \textcolor{blue}{The Moon is shown as a crescent in the top left, waxing crescent, first quarter, waxing gibbous, \textbf{full, waning gibbous}, third quarter, and waning crescent}. Arrows indicate the direction of the waxing and waning of the Moon's phases. The background is solid black, emphasizing the diagram's white and gray elements. There are no background elements or colors other than the white and gray of the Moon and Earth, and the black of the background.\\
\rowcolor{gray!10}
\textbf{Prediction:}  C. waning Gibbous. ( \textcolor{darkgreen}{\faCheck} )\\
\end{tabularx}
\end{tcolorbox}
\caption{\textbf{PrismCaptioner Generates Description to Solve Science \& Technology Task.}}
\label{tab:case5}
\end{table}

\begin{table}[t]
\begin{tcolorbox}[colback=gray!10, colframe=black, arc=5mm, boxrule=0.5mm, width=\textwidth]
\begin{tabularx}{\textwidth}{X}
\rowcolor{gray!10} 
\textbf{Question:} \textbf{For case A accompanying table, answer the questions that follow. Calculate the future value of the annuity, assuming that it is an ordinary annuity.} \\
\rowcolor{gray!10}
\textbf{Image:}\\
\rowcolor{gray!10}
\begin{center}
\includegraphics[width=0.75\textwidth]{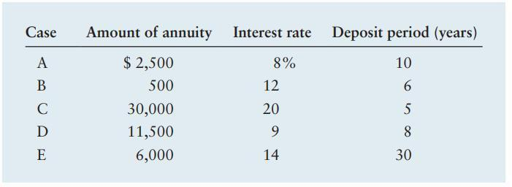}
\end{center} \\
\rowcolor{gray!10}
\textbf{Options:} A. \$126,827 \quad B. \$223,248 \quad C. \$36,216 \quad D. \$4,058 \\
\rowcolor{gray!10}
\textbf{Category:} Math \\
\rowcolor{gray!10}
\textbf{Description:} The image displays a table with five rows and four columns, excluding the header row. The header row is labeled from left to right as "Case," "Amount of annuity," "Interest rate," and "Deposit period (years)." Below the header, there are five cases labeled A, B, C, D, and E. \textcolor{blue}{The "Amount of annuity" column lists monetary values: \$2,500 for case A, \$500 for case B, \$30,000 for case C, \$11,500 for case D, and \$6,000 for case E. The "Interest rate" column lists percentage values: 8\% for case A, 12\% for case B, 20\% for case C, 9\% for case D, and 14\% for case E.} The "Deposit period (years)" column lists numerical values: 10 years for case A, 6 years for case B, 5 years for case C, 8 years for case D, and 30 years for case E. The table is set against a plain background with no additional elements or decorations.\\
\rowcolor{gray!10}
\textbf{Prediction:}  To calculate the future value of an ordinary annuity, you can use the formula:\\
$FV = Pmt * [(1 + r)^n - 1] / r$, \\
\rowcolor{gray!10}
where FV = Future Value, Pmt = Amount of annuity (\$2,500), r = Interest rate (8\% or 0.08), n = Deposit period (10 years).\\
\rowcolor{gray!10}
Plugging in the values:\\
\rowcolor{gray!10}
$FV = \$2,500 * [(1 + 0.08)^{10} - 1] / 0.08$ \\ 
\rowcolor{gray!10}
$FV = \$2,500 * [(1.08)^{10} - 1] / 0.08$ \\
\rowcolor{gray!10}
$FV = \$2,500 * [2.158924 - 1] / 0.08$ \\
\rowcolor{gray!10}
$FV = \$2,500 * 1.158924 / 0.08$ \\
\rowcolor{gray!10}
$FV = \$36,216.55$ \\
\rowcolor{gray!10}
Therefore, the correct answer is: C. \$36,216 ( \textcolor{darkgreen}{\faCheck} ) \\
\end{tabularx}
\end{tcolorbox}
\caption{\textbf{PrismCaptioner Generates Description to Solve Math Task.}}
\label{tab:case6}
\end{table}

\begin{table}[t]
    \centering
    \begin{tabular}{@{}p{0.12\textwidth}p{0.43\textwidth}p{0.38\textwidth}@{}}
        \toprule
        \textbf{Question} & \multicolumn{2}{c}{\textbf{Where is the rug in the living room located?}} \\
        \textbf{} & \multicolumn{2}{c}{\includegraphics[width=0.21\linewidth]{figs/case11.png}} \\
        \textbf{Options} & \multicolumn{2}{c}{A. Next to the door \quad B. Under the table} \\
        \textbf{} & \multicolumn{2}{c}{C. In front of the window \quad D. Under the couch} \\ 
        \midrule
        \textbf{Models} & \multicolumn{1}{c}{\textbf{GPT-4o}} & \multicolumn{1}{c}{\textbf{PrismCaptioner}} \\
        \textbf{Descriptions} & 
        The image depicts a cozy living room corner with a modern and inviting aesthetic. Here is a detailed description of the scene:
        \newline 1. \textbf{Furniture and Objects:}
        \newline Sofa: A gray upholstered sofa is positioned on the left side of the image. It has a few decorative pillows in various patterns and colors, including white, green, and teal.
        \newline Side Table: Next to the sofa, there is a wooden side table with two drawers. On top of the table, there is a blue ceramic lamp with a white lampshade and a potted plant with large green leaves.
        \newline Rug: A gray rug with a white geometric pattern is placed on the floor. The rug is positioned in front of the sofa and extends partially under it, covering the wooden floor.
        \newline Curtains: Light gray curtains with a subtle pattern hang from a black curtain rod above a window on the left side of the image.
        \newline Wall Art: Above the side table, there is a framed piece of art featuring a colorful floral arrangement.
        \newline \textbf{2. Placement and Relationships:}
        \newline \textcolor{blue}{The rug is centrally placed \textbf{in front of the sofa, extending slightly under it}, and covers a portion of the wooden floor.}
        \newline The side table is positioned between the sofa and the door, with the lamp and plant on top providing both functional and decorative elements.
        \newline The framed art is centered above the side table, adding a pop of color to the wall.
        \newline ...
        \newline Overall, the arrangement of furniture and decor items creates a balanced and aesthetically pleasing living room corner. The rug's placement in relation to the sofa and side table helps define the seating area and adds warmth to the space.
        & The image shows a neatly arranged living room corner with a focus on a side table and a portion of a couch. The side table, which appears to be made of wood with a natural finish, has two drawers with metal handles. On top of the table, there is a large blue lamp with a white lampshade, a small white bowl, and a green potted plant with broad leaves. To the left of the side table, there is a couch with a grey upholstery, adorned with a teal throw pillow and a pink throw blanket casually draped over it. \textcolor{red}{The floor is covered with a grey and white patterned area rug, and the walls are painted in a light grey tone.} A window with white trim is partially visible, dressed with patterned curtains in shades of grey and white. On the wall, there is a framed artwork featuring a bouquet of flowers in a vase. The door in the background is a dark grey with a black doorknob, and there is a light switch plate visible on the wall to the right of the door. The overall aesthetic of the room is modern with a touch of warmth provided by the wooden furniture and the cozy textiles.\\
        \midrule
        \textbf{Predictions} & D. Under the couch ( \textcolor{darkgreen}{\faCheck} ) & C. In front of the window ( \textcolor{red}{\faTimes} ) \\
        \bottomrule
    \end{tabular}
    \vspace{2mm}
    \caption{\textbf{Performances between GPT-4o and PrismCaptioner.}}
    \label{tab:case11}
\end{table}

\begin{table}[t]
    \centering
    \begin{tabular}{@{}p{0.13\textwidth}p{0.40\textwidth}p{0.38\textwidth}@{}}
        \toprule
        \textbf{Question} & \multicolumn{2}{c}{\textbf{What is the relative position of the man and the woman sitting at the table?}} \\
        \textbf{} & \multicolumn{2}{c}{\includegraphics[width=0.2\linewidth]{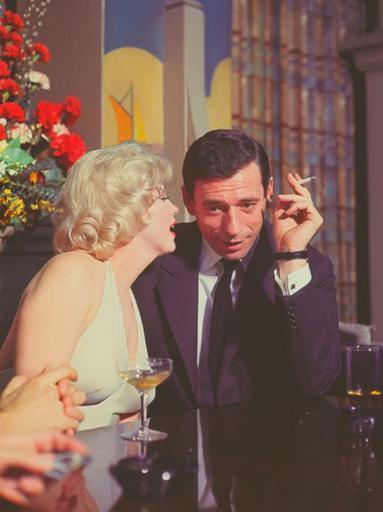}} \\
        \textbf{Options} & \multicolumn{2}{c}{A. The man is on the left and the woman is on the right} \\
        \textbf{} & \multicolumn{2}{c}{B. The man is on the right and the woman is on the left} \\
        \textbf{} & \multicolumn{2}{c}{C. The man and the woman are facing each other} \\
        \textbf{} & \multicolumn{2}{c}{D. The man and the woman are back-to-back} \\
        \midrule
        \textbf{Models} & \multicolumn{1}{c}{\textbf{GPT-4o}} & \multicolumn{1}{c}{\textbf{PrismCaptioner}} \\
        \textbf{Descriptions} & 
        The image depicts a man and a woman sitting closely together at a table in what appears to be a social setting, possibly a bar or a restaurant. \textcolor{blue}{The woman, \textbf{on the left}}, has blonde hair styled in loose curls and is wearing a white dress with a plunging neckline. She is leaning in towards the man, appearing to whisper something in his ear. Her left hand is resting on the table, and her right hand is not visible.
        \newline \textcolor{blue}{The man, \textbf{on the right}}, is wearing a dark suit with a white shirt and a dark tie. He is holding a cigarette in his right hand, which is raised near his face. His left hand is resting on the table. \textcolor{blue}{He is looking directly at the camera with a slight smile.}
        \newline On the table in front of them, there is a champagne glass near the woman and a glass with a dark liquid, possibly a cocktail, near the man. There are also other objects on the table, including what appears to be a hand of another person partially visible on the left side of the image.
        \newline In the background, there is a colorful arrangement of red and white flowers on the left side. The backdrop includes a painted mural or artwork featuring abstract shapes and possibly a cityscape with buildings and a sailboat.
        \newline The spatial arrangement shows the man and woman sitting very close to each other, with the woman leaning in towards the man, indicating a sense of intimacy or a private conversation. \textcolor{blue}{The man is positioned slightly turned towards the camera, while the woman is more focused on him.} & 
        The image depicts a man and a woman seated at a table in what appears to be a bar or restaurant setting. The man is dressed in a dark suit with a white shirt and a dark tie, and he is holding a cigarette between his fingers. The woman is wearing a white dress with a plunging neckline, and her blonde hair is styled in curls. 
        \newline \textcolor{case_ff}{Both individuals are looking at each other}, and the woman seems to be whispering into the man's ear. On the table, there are two glasses, one of which appears to contain a clear liquid, possibly water, and the other is not fully visible. There is also a small, dark object that could be an ashtray or a coaster. 
        \newline In the background, there is a bouquet of flowers with red and white blooms, and the wall behind them features a large, abstract painting with blue and yellow tones. The lighting in the scene is warm, and the overall ambiance suggests a casual yet intimate setting.\\
        \midrule
        \textbf{Predictions} & A. The first image ( \textcolor{darkgreen}{\faCheck} ) & B. The second image ( \textcolor{red}{\faTimes} ) \\
        \bottomrule
    \end{tabular}
    \vspace{2mm}
    \caption{\textbf{Performances between GPT-4o and PrismCaptioner.}}
    \label{tab:case12}
\end{table}

\clearpage

\bibliography{main}

\begin{thebibliography}{10}

\bibitem{2023YiVL}
01-ai.
\newblock Yi-vl.
\newblock https://huggingface.co/01-ai/Yi-VL-34B, 2023.

\bibitem{nocaps}
Harsh Agrawal, Karan Desai, Yufei Wang, Xinlei Chen, Rishabh Jain, Mark
  Johnson, Dhruv Batra, Devi Parikh, Stefan Lee, and Peter Anderson.
\newblock Nocaps: Novel object captioning at scale.
\newblock In {\em Proceedings of the IEEE/CVF international conference on
  computer vision}, pages 8948--8957, 2019.

\bibitem{alayrac2022flamingo}
Jean-Baptiste Alayrac, Jeff Donahue, Pauline Luc, Antoine Miech, Iain Barr,
  Yana Hasson, Karel Lenc, Arthur Mensch, Katherine Millican, Malcolm Reynolds,
  et~al.
\newblock Flamingo: a visual language model for few-shot learning.
\newblock {\em Advances in Neural Information Processing Systems},
  35:23716--23736, 2022.

\bibitem{Claude3}
Anthropic.
\newblock The claude 3 model family: Opus, sonnet, haiku.
\newblock 2024.

\bibitem{qwen}
Jinze Bai, Shuai Bai, Yunfei Chu, Zeyu Cui, Kai Dang, Xiaodong Deng, Yang Fan,
  Wenbin Ge, Yu~Han, Fei Huang, Binyuan Hui, Luo Ji, Mei Li, Junyang Lin, Runji
  Lin, Dayiheng Liu, Gao Liu, Chengqiang Lu, Keming Lu, Jianxin Ma, Rui Men,
  Xingzhang Ren, Xuancheng Ren, Chuanqi Tan, Sinan Tan, Jianhong Tu, Peng Wang,
  Shijie Wang, Wei Wang, Shengguang Wu, Benfeng Xu, Jin Xu, An~Yang, Hao Yang,
  Jian Yang, Shusheng Yang, Yang Yao, Bowen Yu, Hongyi Yuan, Zheng Yuan,
  Jianwei Zhang, Xingxuan Zhang, Yichang Zhang, Zhenru Zhang, Chang Zhou,
  Jingren Zhou, Xiaohuan Zhou, and Tianhang Zhu.
\newblock Qwen technical report.
\newblock {\em arXiv preprint arXiv:2309.16609}, 2023.

\bibitem{bai2023qwen}
Jinze Bai, Shuai Bai, Shusheng Yang, Shijie Wang, Sinan Tan, Peng Wang, Junyang
  Lin, Chang Zhou, and Jingren Zhou.
\newblock Qwen-vl: A frontier large vision-language model with versatile
  abilities.
\newblock {\em arXiv preprint arXiv:2308.12966}, 2023.

\bibitem{cai2024internlm2}
Zheng Cai, Maosong Cao, Haojiong Chen, Kai Chen, Keyu Chen, Xin Chen, Xun Chen,
  Zehui Chen, Zhi Chen, Pei Chu, Xiaoyi Dong, Haodong Duan, Qi~Fan, Zhaoye Fei,
  Yang Gao, Jiaye Ge, Chenya Gu, Yuzhe Gu, Tao Gui, Aijia Guo, Qipeng Guo,
  Conghui He, Yingfan Hu, Ting Huang, Tao Jiang, Penglong Jiao, Zhenjiang Jin,
  Zhikai Lei, Jiaxing Li, Jingwen Li, Linyang Li, Shuaibin Li, Wei Li, Yining
  Li, Hongwei Liu, Jiangning Liu, Jiawei Hong, Kaiwen Liu, Kuikun Liu, Xiaoran
  Liu, Chengqi Lv, Haijun Lv, Kai Lv, Li~Ma, Runyuan Ma, Zerun Ma, Wenchang
  Ning, Linke Ouyang, Jiantao Qiu, Yuan Qu, Fukai Shang, Yunfan Shao, Demin
  Song, Zifan Song, Zhihao Sui, Peng Sun, Yu~Sun, Huanze Tang, Bin Wang,
  Guoteng Wang, Jiaqi Wang, Jiayu Wang, Rui Wang, Yudong Wang, Ziyi Wang,
  Xingjian Wei, Qizhen Weng, Fan Wu, Yingtong Xiong, Chao Xu, Ruiliang Xu, Hang
  Yan, Yirong Yan, Xiaogui Yang, Haochen Ye, Huaiyuan Ying, Jia Yu, Jing Yu,
  Yuhang Zang, Chuyu Zhang, Li~Zhang, Pan Zhang, Peng Zhang, Ruijie Zhang, Shuo
  Zhang, Songyang Zhang, Wenjian Zhang, Wenwei Zhang, Xingcheng Zhang, Xinyue
  Zhang, Hui Zhao, Qian Zhao, Xiaomeng Zhao, Fengzhe Zhou, Zaida Zhou, Jingming
  Zhuo, Yicheng Zou, Xipeng Qiu, Yu~Qiao, and Dahua Lin.
\newblock Internlm2 technical report, 2024.

\bibitem{changpinyo2021conceptual}
Soravit Changpinyo, Piyush Sharma, Nan Ding, and Radu Soricut.
\newblock Conceptual 12m: Pushing web-scale image-text pre-training to
  recognize long-tail visual concepts.
\newblock In {\em Proceedings of the IEEE/CVF conference on computer vision and
  pattern recognition}, pages 3558--3568, 2021.

\bibitem{chen2024allava}
Guiming~Hardy Chen, Shunian Chen, Ruifei Zhang, Junying Chen, Xiangbo Wu, Zhiyi
  Zhang, Zhihong Chen, Jianquan Li, Xiang Wan, and Benyou Wang.
\newblock Allava: Harnessing gpt4v-synthesized data for a lite vision-language
  model, 2024.

\bibitem{chen2024we}
Lin Chen, Jinsong Li, Xiaoyi Dong, Pan Zhang, Yuhang Zang, Zehui Chen, Haodong
  Duan, Jiaqi Wang, Yu~Qiao, Dahua Lin, et~al.
\newblock Are we on the right way for evaluating large vision-language models?
\newblock {\em arXiv preprint arXiv:2403.20330}, 2024.

\bibitem{chen2023sharegpt4v}
Lin Chen, Jisong Li, Xiaoyi Dong, Pan Zhang, Conghui He, Jiaqi Wang, Feng Zhao,
  and Dahua Lin.
\newblock Sharegpt4v: Improving large multi-modal models with better captions.
\newblock {\em arXiv preprint arXiv:2311.12793}, 2023.

\bibitem{chen2024far}
Zhe Chen, Weiyun Wang, Hao Tian, Shenglong Ye, Zhangwei Gao, Erfei Cui, Wenwen
  Tong, Kongzhi Hu, Jiapeng Luo, Zheng Ma, Ji~Ma, Jiaqi Wang, Xiaoyi Dong, Hang
  Yan, Hewei Guo, Conghui He, Botian Shi, Zhenjiang Jin, Chao Xu, Bin Wang,
  Xingjian Wei, Wei Li, Wenjian Zhang, Bo~Zhang, Pinlong Cai, Licheng Wen,
  Xiangchao Yan, Min Dou, Lewei Lu, Xizhou Zhu, Tong Lu, Dahua Lin, Yu~Qiao,
  Jifeng Dai, and Wenhai Wang.
\newblock How far are we to gpt-4v? closing the gap to commercial multimodal
  models with open-source suites, 2024.

\bibitem{chen2023internvl}
Zhe Chen, Jiannan Wu, Wenhai Wang, Weijie Su, Guo Chen, Sen Xing, Muyan Zhong,
  Qinglong Zhang, Xizhou Zhu, Lewei Lu, Bin Li, Ping Luo, Tong Lu, Yu~Qiao, and
  Jifeng Dai.
\newblock Internvl: Scaling up vision foundation models and aligning for
  generic visual-linguistic tasks.
\newblock {\em arXiv preprint arXiv:2312.14238}, 2023.

\bibitem{2023opencompass}
OpenCompass Contributors.
\newblock Opencompass: A universal evaluation platform for foundation models.
\newblock \url{https://github.com/open-compass/opencompass}, 2023.

\bibitem{2023xtuner}
XTuner Contributors.
\newblock Xtuner: A toolkit for efficiently fine-tuning llm.
\newblock \url{https://github.com/InternLM/xtuner}, 2023.

\bibitem{dai2023instructblip}
Wenliang Dai, Junnan Li, Dongxu Li, Anthony Meng~Huat Tiong, Junqi Zhao,
  Weisheng Wang, Boyang Li, Pascale Fung, and Steven Hoi.
\newblock Instructblip: Towards general-purpose vision-language models with
  instruction tuning.
\newblock {\em arXiv preprint arXiv:2305.06500}, 2023.

\bibitem{dettmers2023qlora}
Tim Dettmers, Artidoro Pagnoni, Ari Holtzman, and Luke Zettlemoyer.
\newblock Qlora: Efficient finetuning of quantized llms, 2023.

\bibitem{internlmxcomposer2}
Xiaoyi Dong, Pan Zhang, Yuhang Zang, Yuhang Cao, Bin Wang, Linke Ouyang, Xilin
  Wei, Songyang Zhang, Haodong Duan, Maosong Cao, Wenwei Zhang, Yining Li, Hang
  Yan, Yang Gao, Xinyue Zhang, Wei Li, Jingwen Li, Kai Chen, Conghui He,
  Xingcheng Zhang, Yu~Qiao, Dahua Lin, and Jiaqi Wang.
\newblock Internlm-xcomposer2: Mastering free-form text-image composition and
  comprehension in vision-language large model.
\newblock {\em arXiv preprint arXiv:2401.16420}, 2024.

\bibitem{du2022glm}
Zhengxiao Du, Yujie Qian, Xiao Liu, Ming Ding, Jiezhong Qiu, Zhilin Yang, and
  Jie Tang.
\newblock Glm: General language model pretraining with autoregressive blank
  infilling.
\newblock In {\em Proceedings of the 60th Annual Meeting of the Association for
  Computational Linguistics (Volume 1: Long Papers)}, pages 320--335, 2022.

\bibitem{Fu2023MMEAC}
Chaoyou Fu, Peixian Chen, Yunhang Shen, Yulei Qin, Mengdan Zhang, Xu~Lin,
  Zhenyu Qiu, Wei Lin, Jinrui Yang, Xiawu Zheng, Ke~Li, Xing Sun, and Rongrong
  Ji.
\newblock Mme: A comprehensive evaluation benchmark for multimodal large
  language models.
\newblock {\em ArXiv}, abs/2306.13394, 2023.

\bibitem{fu2024blink}
Xingyu Fu, Yushi Hu, Bangzheng Li, Yu~Feng, Haoyu Wang, Xudong Lin, Dan Roth,
  Noah~A Smith, Wei-Chiu Ma, and Ranjay Krishna.
\newblock Blink: Multimodal large language models can see but not perceive.
\newblock {\em arXiv preprint arXiv:2404.12390}, 2024.

\bibitem{vqav2}
Yash Goyal, Tejas Khot, Douglas Summers-Stay, Dhruv Batra, and Devi Parikh.
\newblock Making the v in vqa matter: Elevating the role of image understanding
  in visual question answering.
\newblock In {\em Proceedings of the IEEE conference on computer vision and
  pattern recognition}, pages 6904--6913, 2017.

\bibitem{han2023infimm}
Xiaotian Han, Quanzeng You, Yongfei Liu, Wentao Chen, Huangjie Zheng, Khalil
  Mrini, Xudong Lin, Yiqi Wang, Bohan Zhai, Jianbo Yuan, et~al.
\newblock Infimm-eval: Complex open-ended reasoning evaluation for multi-modal
  large language models.
\newblock {\em arXiv e-prints}, pages arXiv--2311, 2023.

\bibitem{hudson2019gqa}
Drew~A Hudson and Christopher~D Manning.
\newblock Gqa: A new dataset for real-world visual reasoning and compositional
  question answering.
\newblock In {\em Proceedings of the IEEE/CVF conference on computer vision and
  pattern recognition}, pages 6700--6709, 2019.

\bibitem{kembhavi2016diagram}
Aniruddha Kembhavi, Mike Salvato, Eric Kolve, Minjoon Seo, Hannaneh Hajishirzi,
  and Ali Farhadi.
\newblock A diagram is worth a dozen images.
\newblock In {\em Computer Vision--ECCV 2016: 14th European Conference,
  Amsterdam, The Netherlands, October 11--14, 2016, Proceedings, Part IV 14},
  pages 235--251. Springer, 2016.

\bibitem{laurencon2023obelics}
Hugo Laurençon, Lucile Saulnier, Léo Tronchon, Stas Bekman, Amanpreet Singh,
  Anton Lozhkov, Thomas Wang, Siddharth Karamcheti, Alexander~M. Rush, Douwe
  Kiela, Matthieu Cord, and Victor Sanh.
\newblock Obelics: An open web-scale filtered dataset of interleaved image-text
  documents, 2023.

\bibitem{li2023seed}
Bohao Li, Rui Wang, Guangzhi Wang, Yuying Ge, Yixiao Ge, and Ying Shan.
\newblock Seed-bench: Benchmarking multimodal llms with generative
  comprehension.
\newblock {\em arXiv preprint arXiv:2307.16125}, 2023.

\bibitem{li2023blip}
Junnan Li, Dongxu Li, Silvio Savarese, and Steven Hoi.
\newblock Blip-2: Bootstrapping language-image pre-training with frozen image
  encoders and large language models.
\newblock {\em arXiv preprint arXiv:2301.12597}, 2023.

\bibitem{li2023evaluating}
Yifan Li, Yifan Du, Kun Zhou, Jinpeng Wang, Wayne~Xin Zhao, and Ji-Rong Wen.
\newblock Evaluating object hallucination in large vision-language models.
\newblock {\em arXiv preprint arXiv:2305.10355}, 2023.

\bibitem{lin2014microsoft}
Tsung-Yi Lin, Michael Maire, Serge Belongie, James Hays, Pietro Perona, Deva
  Ramanan, Piotr Doll{\'a}r, and C~Lawrence Zitnick.
\newblock Microsoft coco: Common objects in context.
\newblock In {\em Computer Vision--ECCV 2014: 13th European Conference, Zurich,
  Switzerland, September 6-12, 2014, Proceedings, Part V 13}, pages 740--755.
  Springer, 2014.

\bibitem{Liu2022VisualSR}
Fangyu Liu, Guy Edward~Toh Emerson, and Nigel Collier.
\newblock Visual spatial reasoning.
\newblock {\em Transactions of the Association for Computational Linguistics},
  2023.

\bibitem{liu2023hallusionbench}
Fuxiao Liu, Tianrui Guan, Zongxia Li, Lichang Chen, Yaser Yacoob, Dinesh
  Manocha, and Tianyi Zhou.
\newblock Hallusionbench: You see what you think? or you think what you see? an
  image-context reasoning benchmark challenging for gpt-4v (ision), llava-1.5,
  and other multi-modality models.
\newblock {\em arXiv preprint arXiv:2310.14566}, 2023.

\bibitem{liu2023improved}
Haotian Liu, Chunyuan Li, Yuheng Li, and Yong~Jae Lee.
\newblock Improved baselines with visual instruction tuning.
\newblock {\em arXiv preprint arXiv:2310.03744}, 2023.

\bibitem{liu2024llavanext}
Haotian Liu, Chunyuan Li, Yuheng Li, Bo~Li, Yuanhan Zhang, Sheng Shen, and
  Yong~Jae Lee.
\newblock Llava-next: Improved reasoning, ocr, and world knowledge, January
  2024.

\bibitem{liu2023visual}
Haotian Liu, Chunyuan Li, Qingyang Wu, and Yong~Jae Lee.
\newblock Visual instruction tuning.
\newblock {\em arXiv preprint arXiv:2304.08485}, 2023.

\bibitem{liu2023mmbench}
Yuan Liu, Haodong Duan, Yuanhan Zhang, Bo~Li, Songyang Zhang, Wangbo Zhao, Yike
  Yuan, Jiaqi Wang, Conghui He, Ziwei Liu, et~al.
\newblock Mmbench: Is your multi-modal model an all-around player?
\newblock {\em arXiv preprint arXiv:2307.06281}, 2023.

\bibitem{liu2023hidden}
Yuliang Liu, Zhang Li, Hongliang Li, Wenwen Yu, Mingxin Huang, Dezhi Peng,
  Mingyu Liu, Mingrui Chen, Chunyuan Li, Lianwen Jin, et~al.
\newblock On the hidden mystery of ocr in large multimodal models.
\newblock {\em arXiv preprint arXiv:2305.07895}, 2023.

\bibitem{lu2024deepseek}
Haoyu Lu, Wen Liu, Bo~Zhang, Bingxuan Wang, Kai Dong, Bo~Liu, Jingxiang Sun,
  Tongzheng Ren, Zhuoshu Li, Yaofeng Sun, et~al.
\newblock Deepseek-vl: towards real-world vision-language understanding.
\newblock {\em arXiv preprint arXiv:2403.05525}, 2024.

\bibitem{lu2023mathvista}
Pan Lu, Hritik Bansal, Tony Xia, Jiacheng Liu, Chunyuan Li, Hannaneh
  Hajishirzi, Hao Cheng, Kai-Wei Chang, Michel Galley, and Jianfeng Gao.
\newblock Mathvista: Evaluating mathematical reasoning of foundation models in
  visual contexts.
\newblock {\em arXiv preprint arXiv:2310.02255}, 2023.

\bibitem{ok-vqa}
Kenneth Marino, Mohammad Rastegari, Ali Farhadi, and Roozbeh Mottaghi.
\newblock Ok-vqa: A visual question answering benchmark requiring external
  knowledge.
\newblock In {\em Proceedings of the IEEE/cvf conference on computer vision and
  pattern recognition}, pages 3195--3204, 2019.

\bibitem{masry2022chartqa}
Ahmed Masry, Do~Xuan Long, Jia~Qing Tan, Shafiq Joty, and Enamul Hoque.
\newblock Chartqa: A benchmark for question answering about charts with visual
  and logical reasoning.
\newblock {\em arXiv preprint arXiv:2203.10244}, 2022.

\bibitem{mathew2021docvqa}
Minesh Mathew, Dimosthenis Karatzas, and CV~Jawahar.
\newblock Docvqa: A dataset for vqa on document images.
\newblock In {\em Proceedings of the IEEE/CVF winter conference on applications
  of computer vision}, pages 2200--2209, 2021.

\bibitem{mishra2019ocr}
Anand Mishra, Shashank Shekhar, Ajeet~Kumar Singh, and Anirban Chakraborty.
\newblock Ocr-vqa: Visual question answering by reading text in images.
\newblock In {\em 2019 international conference on document analysis and
  recognition (ICDAR)}, pages 947--952. IEEE, 2019.

\bibitem{2022chatgpt}
OpenAI.
\newblock Chatgpt.
\newblock \url{https://openai.com/blog/chatgpt}, 2023.

\bibitem{OpenAI2023GPT4TR}
OpenAI.
\newblock Gpt-4 technical report.
\newblock {\em ArXiv}, abs/2303.08774, 2023.

\bibitem{minicpm2024}
OpenBMB.
\newblock Minicpm: Unveiling the potential of end-side large language models,
  2024.

\bibitem{ormazabal2024reka}
Aitor Ormazabal, Che Zheng, Cyprien de~Masson d'Autume, Dani Yogatama, Deyu Fu,
  Donovan Ong, Eric Chen, Eugenie Lamprecht, Hai Pham, Isaac Ong, et~al.
\newblock Reka core, flash, and edge: A series of powerful multimodal language
  models.
\newblock {\em arXiv preprint arXiv:2404.12387}, 2024.

\bibitem{peng2023kosmos}
Zhiliang Peng, Wenhui Wang, Li~Dong, Yaru Hao, Shaohan Huang, Shuming Ma, and
  Furu Wei.
\newblock Kosmos-2: Grounding multimodal large language models to the world.
\newblock {\em arXiv preprint arXiv:2306.14824}, 2023.

\bibitem{radford2021learning}
Alec Radford, Jong~Wook Kim, Chris Hallacy, Aditya Ramesh, Gabriel Goh,
  Sandhini Agarwal, Girish Sastry, Amanda Askell, Pamela Mishkin, Jack Clark,
  et~al.
\newblock Learning transferable visual models from natural language
  supervision.
\newblock In {\em International conference on machine learning}, pages
  8748--8763. PMLR, 2021.

\bibitem{schuhmann2021laion}
Christoph Schuhmann, Richard Vencu, Romain Beaumont, Robert Kaczmarczyk,
  Clayton Mullis, Aarush Katta, Theo Coombes, Jenia Jitsev, and Aran
  Komatsuzaki.
\newblock Laion-400m: Open dataset of clip-filtered 400 million image-text
  pairs.
\newblock {\em arXiv preprint arXiv:2111.02114}, 2021.

\bibitem{textvqa}
Amanpreet Singh, Vivek Natarajan, Meet Shah, Yu~Jiang, Xinlei Chen, Dhruv
  Batra, Devi Parikh, and Marcus Rohrbach.
\newblock Towards vqa models that can read.
\newblock In {\em Proceedings of the IEEE/CVF conference on computer vision and
  pattern recognition}, pages 8317--8326, 2019.

\bibitem{sun2023emu2chat}
Quan Sun, Yufeng Cui, Xiaosong Zhang, Fan Zhang, Qiying Yu, Zhengxiong Luo,
  Yueze Wang, Yongming Rao, Jingjing Liu, Tiejun Huang, et~al.
\newblock Generative multimodal models are in-context learners.
\newblock {\em arXiv preprint arXiv:2312.13286}, 2023.

\bibitem{sun2023eva}
Quan Sun, Yuxin Fang, Ledell Wu, Xinlong Wang, and Yue Cao.
\newblock Eva-clip: Improved training techniques for clip at scale.
\newblock {\em arXiv preprint arXiv:2303.15389}, 2023.

\bibitem{team2023gemini}
Gemini Team, Rohan Anil, Sebastian Borgeaud, Yonghui Wu, Jean-Baptiste Alayrac,
  Jiahui Yu, Radu Soricut, Johan Schalkwyk, Andrew~M Dai, Anja Hauth, et~al.
\newblock Gemini: a family of highly capable multimodal models.
\newblock {\em arXiv preprint arXiv:2312.11805}, 2023.

\bibitem{2023internlm}
InternLM Team.
\newblock Internlm: A multilingual language model with progressively enhanced
  capabilities.
\newblock \url{https://github.com/InternLM/InternLM-techreport}, 2023.

\bibitem{tong2024eyes}
Shengbang Tong, Zhuang Liu, Yuexiang Zhai, Yi~Ma, Yann LeCun, and Saining Xie.
\newblock Eyes wide shut? exploring the visual shortcomings of multimodal llms.
\newblock {\em arXiv preprint arXiv:2401.06209}, 2024.

\bibitem{touvron2023llama}
Hugo Touvron, Thibaut Lavril, Gautier Izacard, Xavier Martinet, Marie-Anne
  Lachaux, Timoth{\'e}e Lacroix, Baptiste Rozi{\`e}re, Naman Goyal, Eric
  Hambro, Faisal Azhar, et~al.
\newblock Llama: Open and efficient foundation language models.
\newblock {\em arXiv preprint arXiv:2302.13971}, 2023.

\bibitem{touvron2023llama2}
Hugo Touvron, Louis Martin, Kevin Stone, Peter Albert, Amjad Almahairi, Yasmine
  Babaei, Nikolay Bashlykov, Soumya Batra, Prajjwal Bhargava, Shruti Bhosale,
  et~al.
\newblock Llama 2: Open foundation and fine-tuned chat models.
\newblock {\em arXiv preprint arXiv:2307.09288}, 2023.

\bibitem{wei2023chainofthought}
Jason Wei, Xuezhi Wang, Dale Schuurmans, Maarten Bosma, Brian Ichter, Fei Xia,
  Ed~Chi, Quoc Le, and Denny Zhou.
\newblock Chain-of-thought prompting elicits reasoning in large language
  models, 2023.

\bibitem{RealWorldQA}
XAI.
\newblock Grok-1.5 vision preview.
\newblock 2024.

\bibitem{Xu2023LVLMeHubAC}
Peng Xu, Wenqi Shao, Kaipeng Zhang, Peng Gao, Shuo Liu, Meng Lei, Fanqing Meng,
  Siyuan Huang, Yu~Jiao Qiao, and Ping Luo.
\newblock Lvlm-ehub: A comprehensive evaluation benchmark for large
  vision-language models.
\newblock 2023.

\bibitem{ye2023mplug}
Qinghao Ye, Haiyang Xu, Guohai Xu, Jiabo Ye, Ming Yan, Yiyang Zhou, Junyang
  Wang, Anwen Hu, Pengcheng Shi, Yaya Shi, et~al.
\newblock mplug-owl: Modularization empowers large language models with
  multimodality.
\newblock {\em arXiv preprint arXiv:2304.14178}, 2023.

\bibitem{ye2023mplugowl2}
Qinghao Ye, Haiyang Xu, Jiabo Ye, Ming Yan, Anwen Hu, Haowei Liu, Qi~Qian,
  Ji~Zhang, Fei Huang, and Jingren Zhou.
\newblock mplug-owl2: Revolutionizing multi-modal large language model with
  modality collaboration, 2023.

\bibitem{yue2023mmmu}
Xiang Yue, Yuansheng Ni, Kai Zhang, Tianyu Zheng, Ruoqi Liu, Ge~Zhang, Samuel
  Stevens, Dongfu Jiang, Weiming Ren, Yuxuan Sun, et~al.
\newblock Mmmu: A massive multi-discipline multimodal understanding and
  reasoning benchmark for expert agi.
\newblock {\em arXiv preprint arXiv:2311.16502}, 2023.

\bibitem{zhai2023sigmoid}
Xiaohua Zhai, Basil Mustafa, Alexander Kolesnikov, and Lucas Beyer.
\newblock Sigmoid loss for language image pre-training.
\newblock In {\em Proceedings of the IEEE/CVF International Conference on
  Computer Vision}, pages 11975--11986, 2023.

\bibitem{zhang2024far}
Yizhe Zhang, He~Bai, Ruixiang Zhang, Jiatao Gu, Shuangfei Zhai, Josh Susskind,
  and Navdeep Jaitly.
\newblock How far are we from intelligent visual deductive reasoning?
\newblock {\em arXiv preprint arXiv:2403.04732}, 2024.

\bibitem{zhu2023minigpt}
Deyao Zhu, Jun Chen, Xiaoqian Shen, Xiang Li, and Mohamed Elhoseiny.
\newblock Minigpt-4: Enhancing vision-language understanding with advanced
  large language models.
\newblock {\em arXiv preprint arXiv:2304.10592}, 2023.

\end{thebibliography}
\bibliographystyle{plain}

\end{document}